%% file: main.tex
\documentclass{article}

\usepackage{standalone}
\usepackage{microtype}
\usepackage{graphicx}
\usepackage{subcaption}
\usepackage{booktabs} 
\usepackage{nicefrac}
\usepackage{enumitem}

\usepackage{hyperref}

\usepackage[preprint]{icml2026}

\usepackage{amsmath}
\usepackage{amssymb}
\usepackage{mathtools}
\usepackage{amsthm}
\usepackage{dsfont}

\DeclareMathOperator{\sample}{sample}

\usepackage{etoolbox}
\apptocmd{\normalsize}{%
  \setlength{\abovedisplayskip}{10pt}%
  \setlength{\belowdisplayskip}{10pt}%
  \setlength{\abovedisplayshortskip}{0pt}%
  \setlength{\belowdisplayshortskip}{3pt}%
}{}{}
\usepackage{float}
\usepackage{changepage}
\usepackage{graphicx}
\definecolor{forrest}{RGB}{76,175,80}

\usepackage[capitalize,noabbrev]{cleveref}

\theoremstyle{plain}

\theoremstyle{definition}

\theoremstyle{remark}

\usepackage{tikz}
\usepackage{tikz-3dplot}
\usetikzlibrary{arrows.meta}
\usetikzlibrary{decorations.markings, arrows.meta}

\usepackage[textsize=tiny]{todonotes}

\icmltitlerunning{Sampling error and internal states --- A geometric perspective}

\begin{document}
\twocolumn[
  \icmltitle{A geometric relation of the error introduced by sampling a language model's output distribution to its internal state}



  \icmlsetsymbol{equal}{*}

  \begin{icmlauthorlist}
    \icmlauthor{Albert F. Modenbach}{aaa}
  \end{icmlauthorlist}

  \icmlaffiliation{aaa}{Department of Mathematics, King's College London, United Kingdom}

  \icmlcorrespondingauthor{Albert F. Modenbach}{a\_modenbach@outlook.com}

  \icmlkeywords{Machine Learning, ICML}

  \vskip 0.3in
]



\printAffiliationsAndNotice{}  

\begin{abstract}
  GPT-style language models are sensitive to single-token changes at generation points where the predicted probability distribution is spread across multiple tokens. Viewing this sensitivity as a geometric property, we derive an $\mathfrak{so}(n)$-valued 1-form that depends only on the geometry of the token embeddings. Despite this purely geometric origin, we show that its curvature is semantically meaningful: On chess reasoning tasks, the curvature couples to the world model of an off-the-shelf instruction-tuned model, with transformations clustering by board region and respecting piece importance. Our findings suggest that token space geometry directly reflects how models internally represent problems\footnote{Code to reproduce all experiments is provided in the supplementary material.}.
\end{abstract}

\input{body.tex}

\section*{Impact Statement}
This paper presents work whose goal is to advance the field of Machine
Learning. There are many potential societal consequences of our work, none
which we feel must be specifically highlighted here.

\bibliography{main}
\bibliographystyle{icml2026}
\newpage
\onecolumn
\section*{Appendix A}
\begin{figure}[h]
    \centering 
    \includegraphics[width=0.3\textwidth]{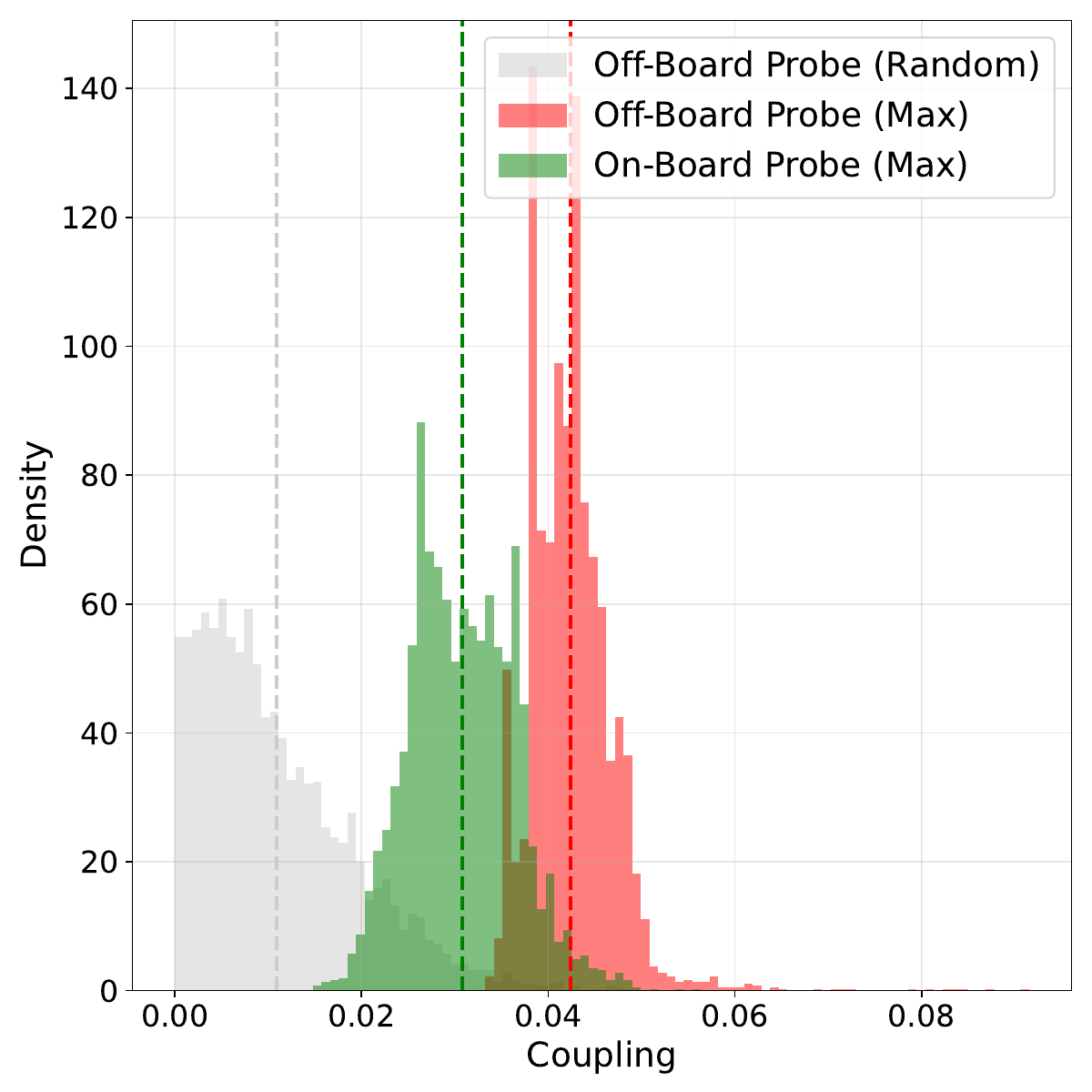}
    \includegraphics[width=0.3\textwidth]{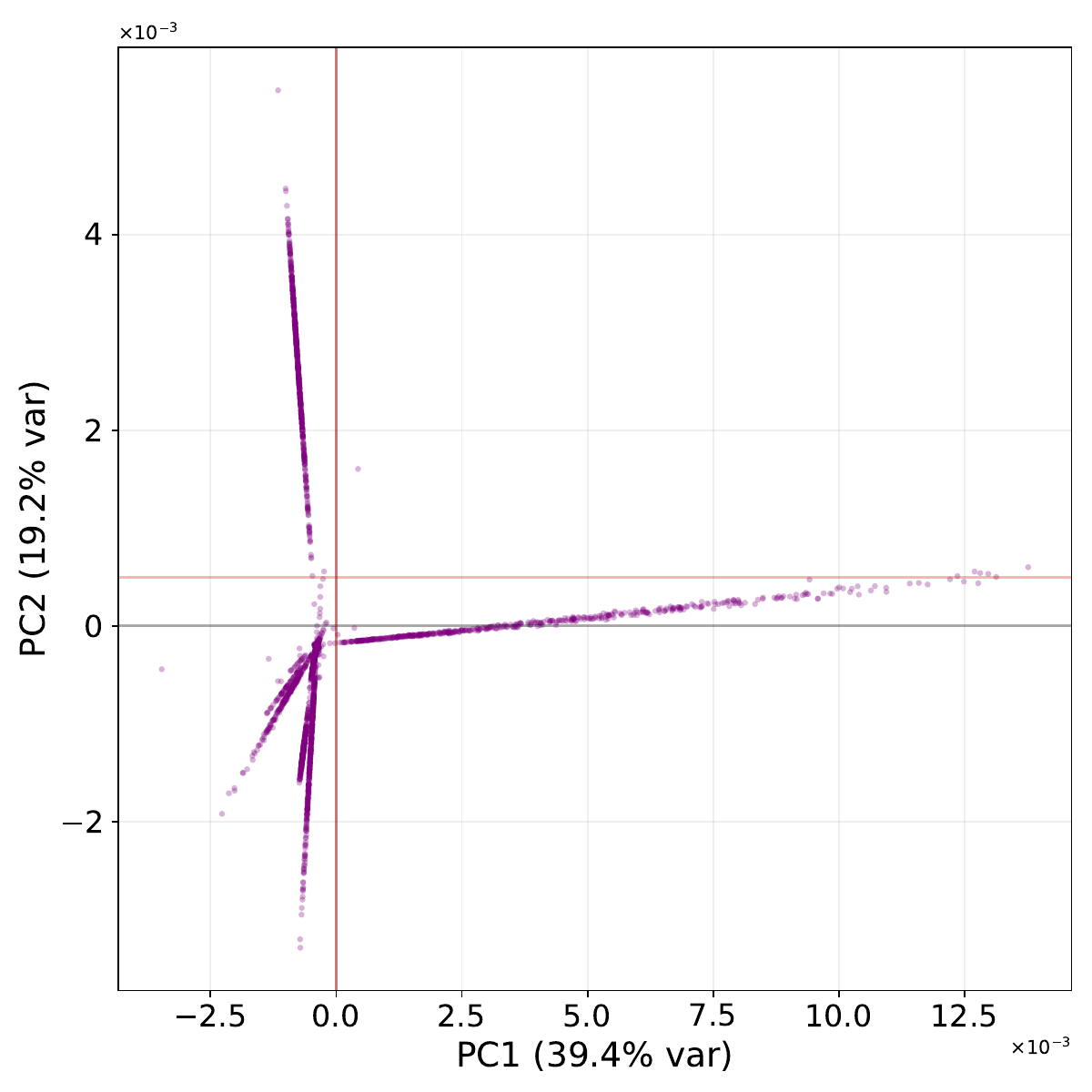}
    \includegraphics[width=0.3\textwidth]{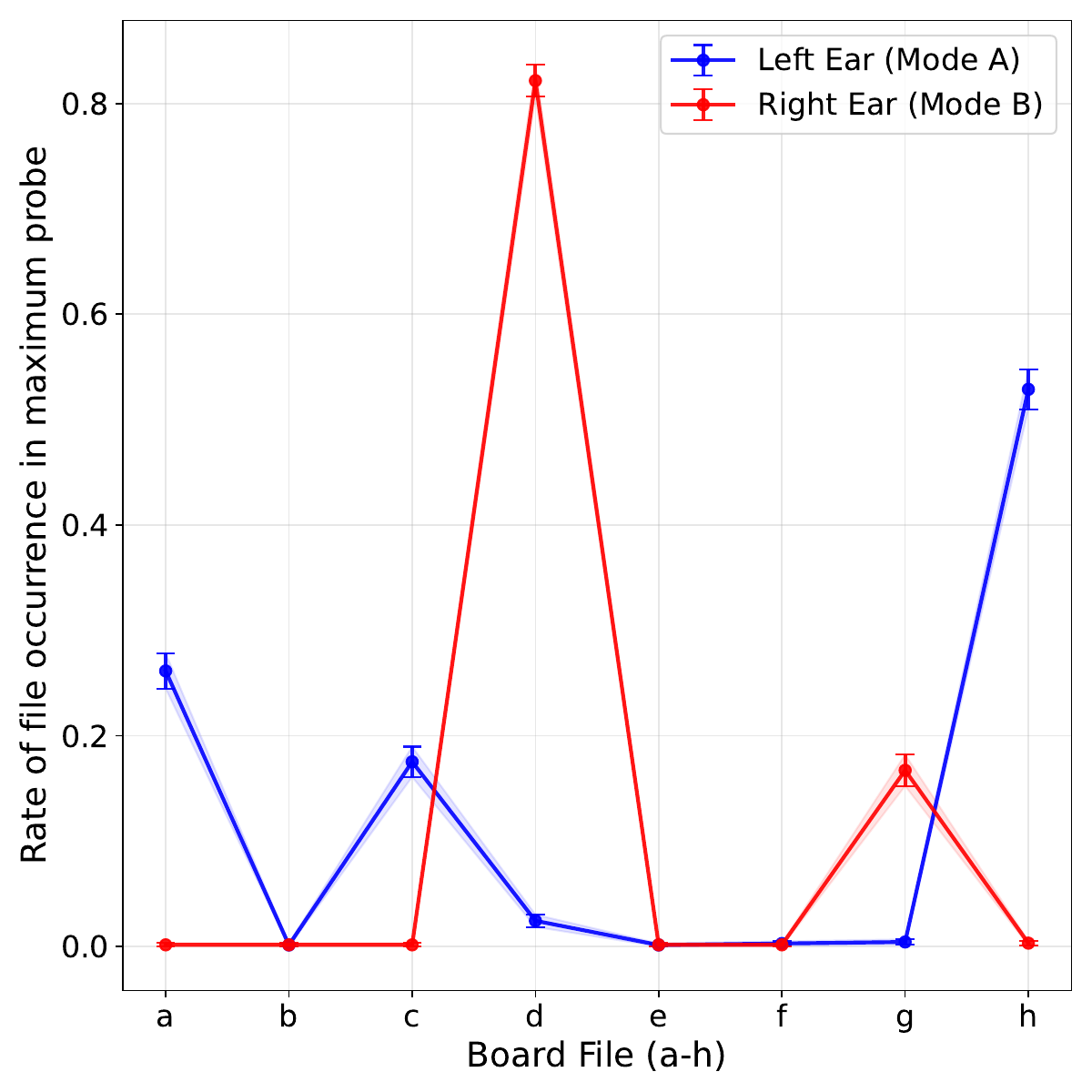}
    \includegraphics[width=0.3\textwidth]{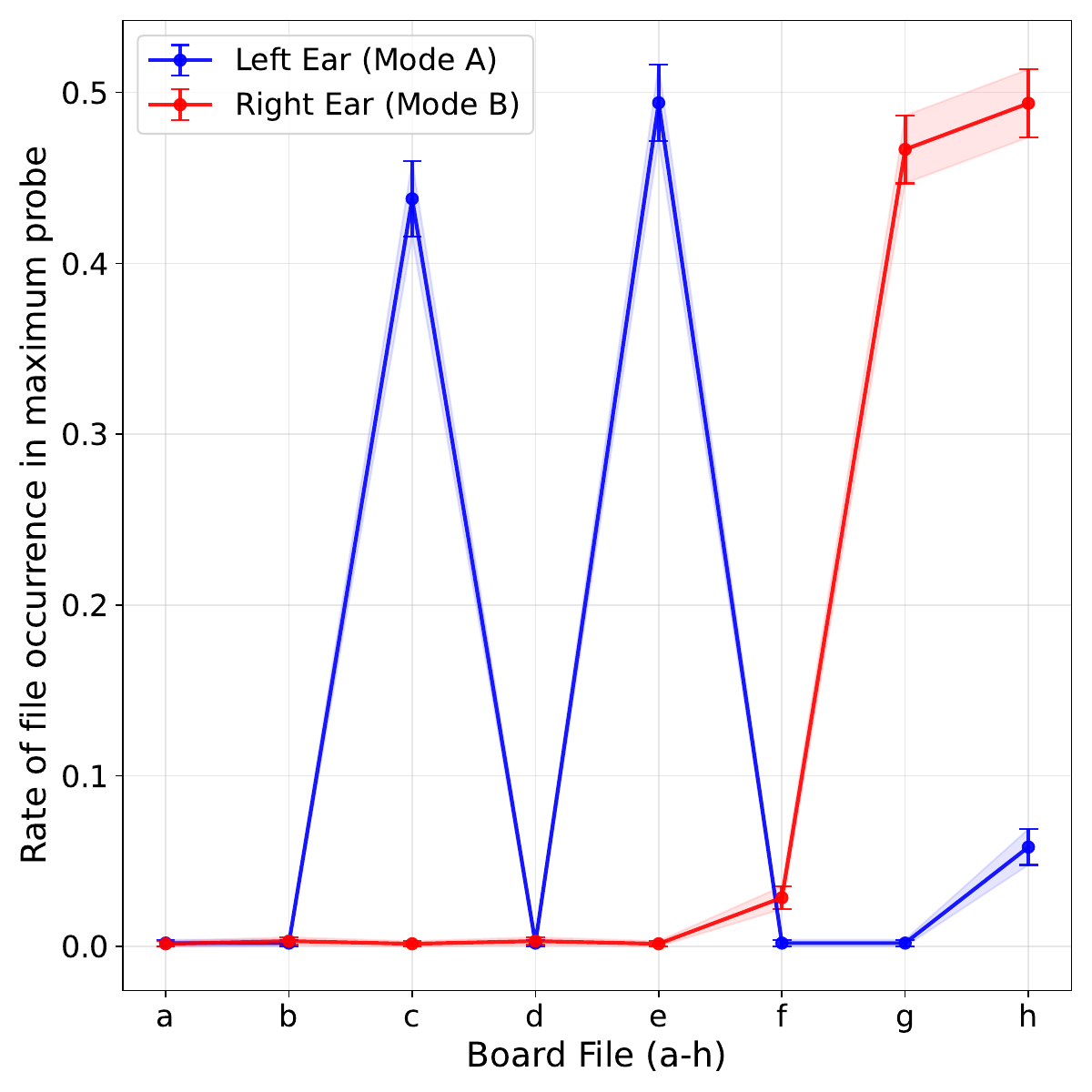}
    \includegraphics[width=0.3\textwidth]{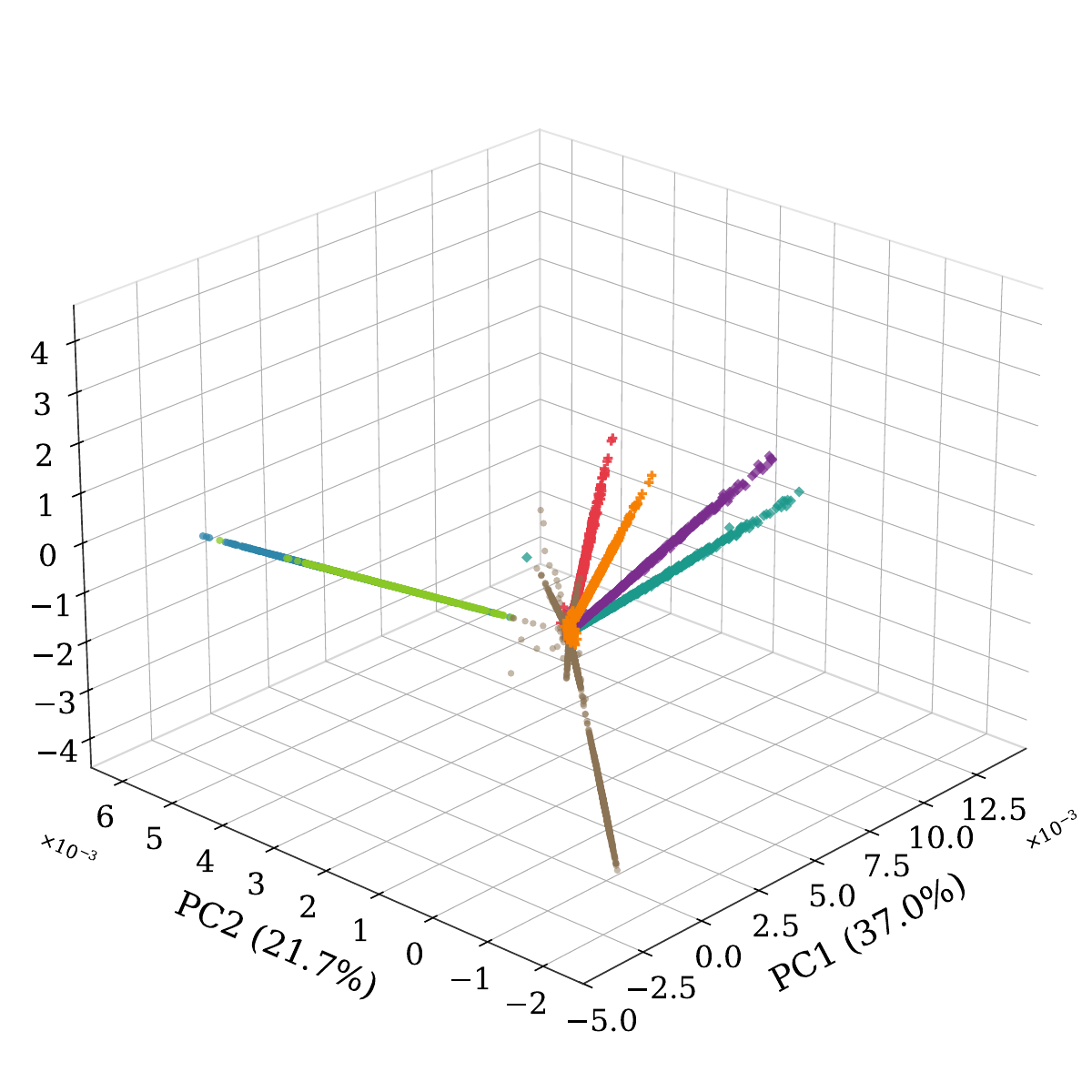}
    \includegraphics[width=0.3\textwidth]{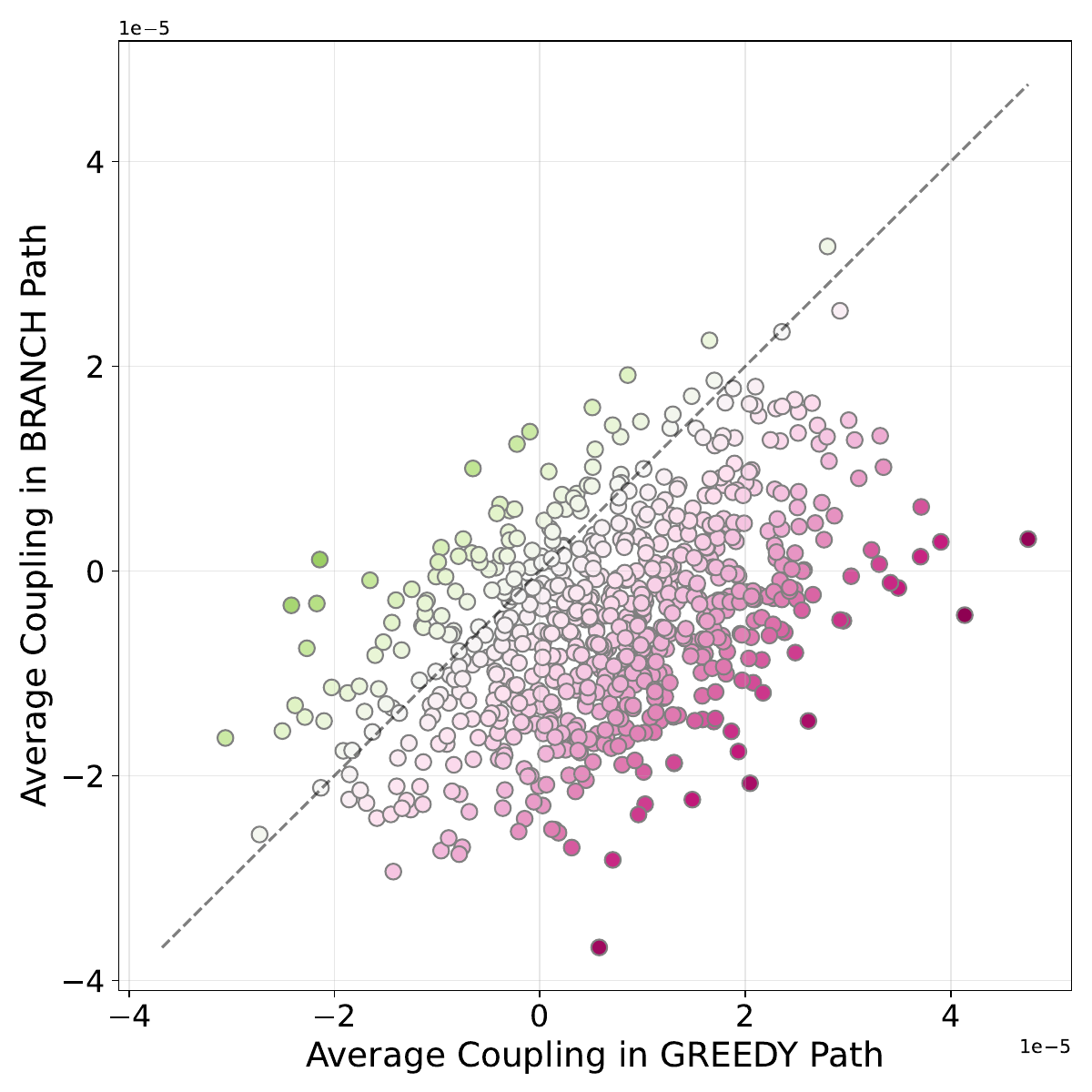}
    \includegraphics[width=0.3\textwidth]{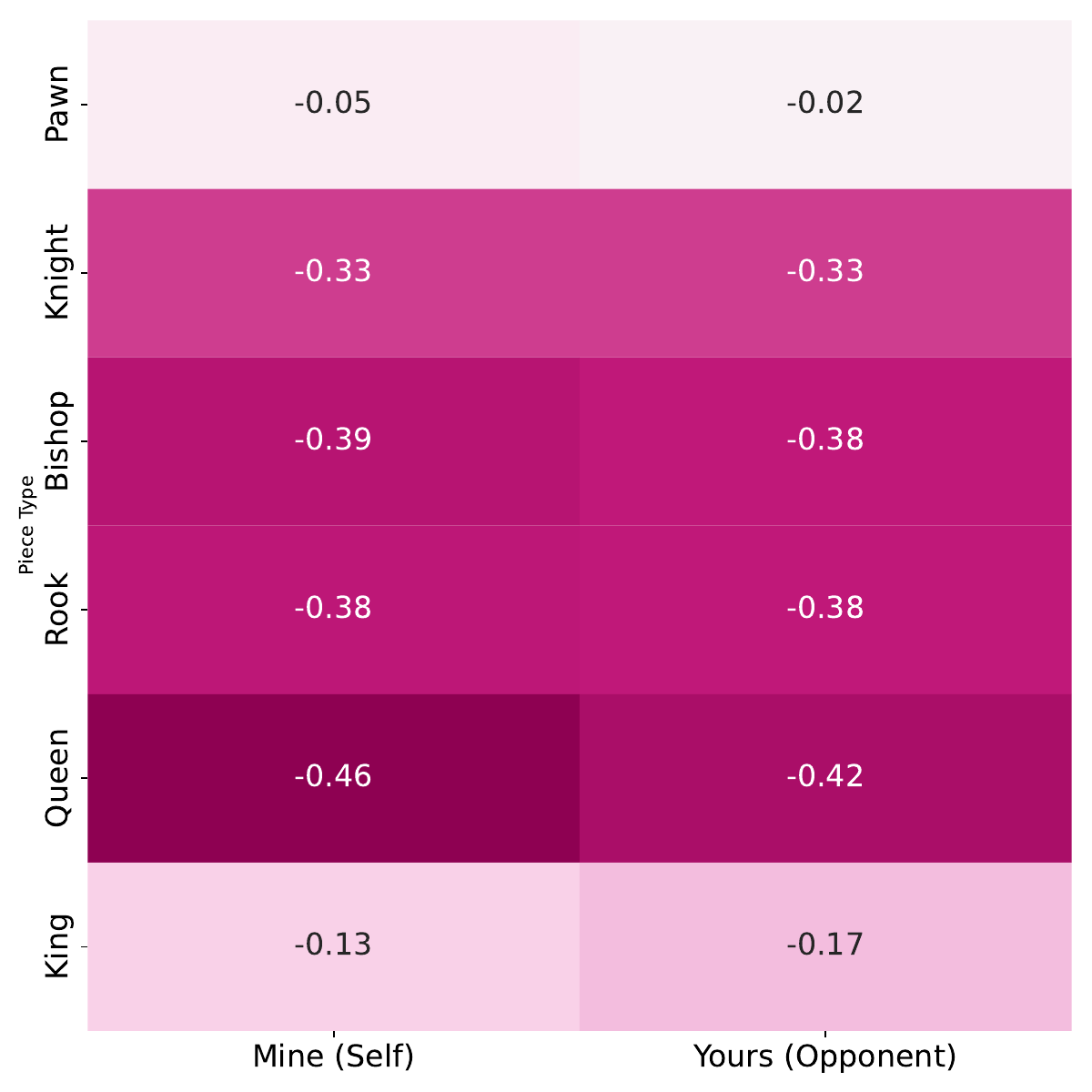}
    \includegraphics[width=0.3\textwidth]{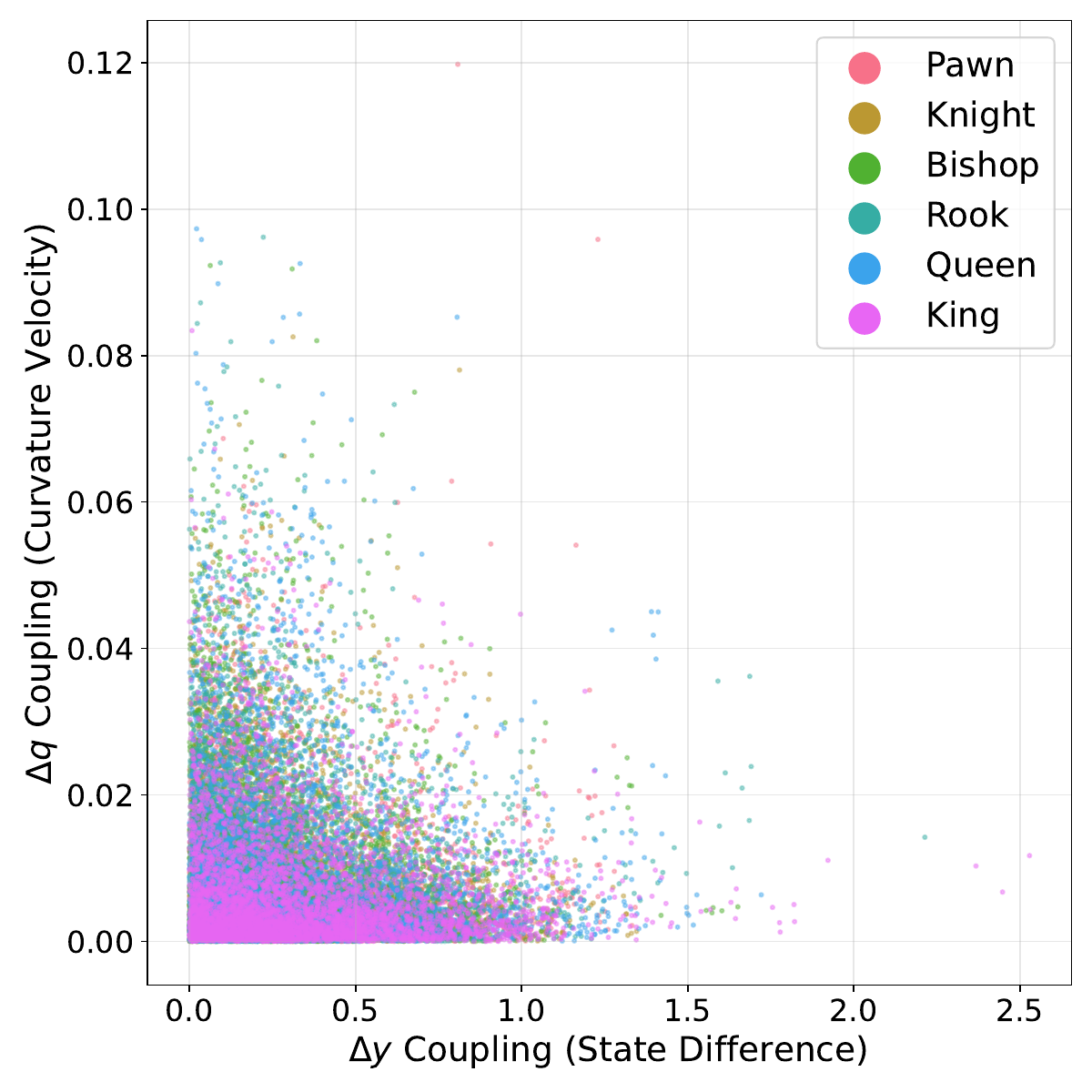}
    \caption{Collection of results graph like those in figure \ref{fig:probe_seperation_and_pca}, \ref{fig:chess-files}, \ref{fig:energy_and_semantic_shift}, and \ref{fig:pca-three-d_and_spectrum} for the Mistral model.}
    \label{fig:mistral_results}
\end{figure}
\newpage
\section*{Appendix B}
\begin{figure}[h!]
  \centering
  \includegraphics*[width=0.7\textwidth]{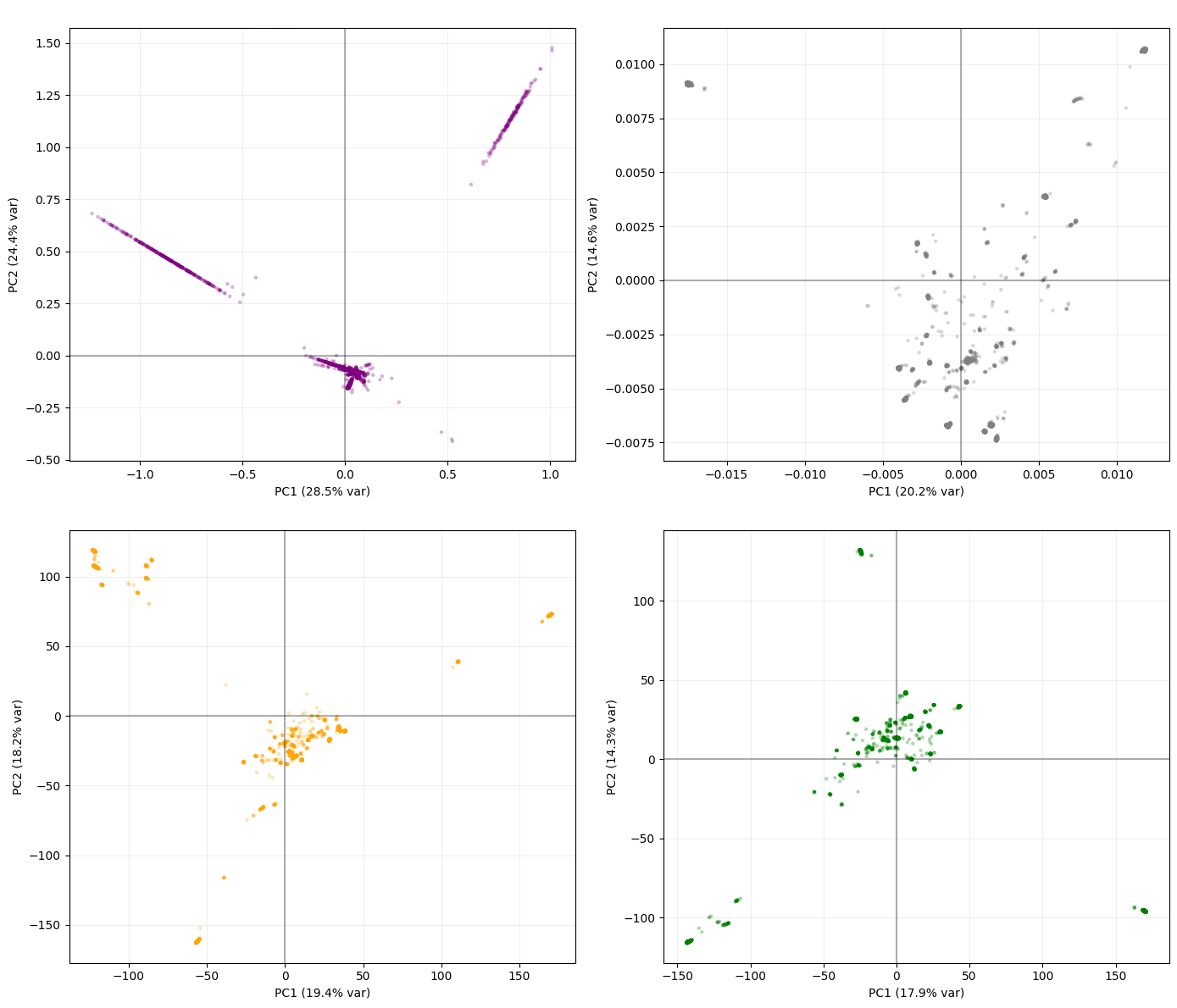}
  \caption{PCA projections of $q = Hy - y$ for the true holonomy operator, versus a scale-matched random SO(n) matrix, and a holonomy operator rotating $z_t$ onto $v_1$ and one onto $v_2$. The structured V-pattern and higher variance concentration ($52.9\%$ vs $34.5\%$ and $31.4\%$) in the true holonomy confirms that the semantic clustering arises from the specific geometry of token-space blurring, not arbitrary rotations.}
  \label{fig:apx:holonomy_ablation}
\end{figure}

\end{document}

%% file: body.tex
\section{Introduction}

One can generate simple images by training a feed-forward neural network to map one-hot encoded labels to a $w\times h$: $w, h\in\mathbb N$ sized vector of pixel brightness values. It is well known that such a simple architecture's reproductions will only ever be blurry, as the best solution to the given mean-square-error optimization problem is to reproduce the average image associated with each label.

Let $\mathcal Z\cong\mathbb R^n$ be the vector space of a GPT-style \cite{radford2018improving} language model's hidden states. Where $n$ is the model's hidden dimension, and $z \in \mathcal Z$ is the last hidden activation of the forward pass. We wish to write the forward pass at a particular time point as $F:\mathcal Z\to\mathcal Z$, and so state it slightly unusually as
\begin{align*}
    z_{t+1} = F(W[\sample(\sigma(Vz_t))]).  
\end{align*}
Here $V \in \mathbb R^{l\times n}$ and  $W \in \mathbb R^{l\times n}$ are the un-embedding and embedding vocabulary matrices respectively, that take a vector in the hidden activation space to the vocabulary space (or vice-versa), $\mathcal V\cong \mathbb R^l$, with vocabulary size $l$. $\sigma$ is the usual softmax function with a fixed temperature of $1$. The $W[\cdot]$ notation is $W[i] = e_i^TW$, so selecting the $i$-th row from the vocabulary matrix, and \textit{sample} is the function used to select an index from the $\sigma(Vz_t)$ probability distribution.

$F$ depends on the history of the token sequence leading up to a time $t$, but we will treat this implicitly to lighten notation.

Now, for some generation points, $z_t$ might be oriented in such a way that it is not well aligned to any of the token embedding vectors, and so gives rise to a probability distribution with significant mass on the top two tokens. By analogy to the image generating feed forward network, we say that these points have become \textit{blurred}: The model is producing an ``averaged'' output between the top two tokens. At these points there is no good choice \texttt{sample} can make and --- because the input into the model at the next time step is the selected token's embedding vector --- some of the information that $z_t$ was capturing must have been destroyed by the token selection.

Let $v_1\in\mathcal Z$ and $v_2\in\mathcal Z$ be the top and second best token embedding vectors as determined by the sampling process at a time $t$:
\begin{align}
    \label{eqn:vonevtwo}
    v_1 &= W[\text{top}(\sigma(Vz_t))],\nonumber\\
    v_2 &= W[\text{second}(\sigma(Vz_t))].
\end{align}
In this paper, we are going to develop how the volume of the parallelepiped spanned by the vectors $z_t$, $(p_1v_1)$, and $(p_2v_2)$ (where $p_i$ are the probability masses) measures the extent of blurring, and how the wedge product\footnote{Wedge products are in some sense a generalisation of the cross-product to higher dimension. A wedge between two vectors in three dimensional space $A \wedge B\equiv A\times B$ is exactly the cross-product. Through this, wedge products also generalise the notion of volumes and areas, so that $||A\wedge B \wedge C||$ is the volume of the parallelepiped spanned by these three vectors. Note that the ordinary cross-product is only well defined in three dimensions.} between them leads to an oriented measure of uncertainty. We postulate that the geometry of blurring relates the model's internal state to its learned world model, and show that this product defines one such relationship. We also show that similar but much simpler transformations do not successfully relate the internal state to the learned world model.

\section{Nomenclature and Experimental Notes}
\label{sec:nomenclature_and_notes}
Before deriving our geometric relationship, let us outline the experimental context we are working in.

All our experiments were performed using the \texttt{\small Qwen2.5-32B-Instruct} \cite{qwen2.5} and \texttt{\small Mistral-Small-3.1-24B-Instruct-2503} \cite{mistralsmall31} models. We developed everything using the Qwen model first, only evaluating the Mistral model after finalizing our claims to reserve it as a held out model.

Beyond the definitions made in the introduction, we will treat $y\in\mathcal Y\cong\mathbb R^n$ as the vector in the residual stream at layer $20$ and $30$ of the Qwen and Mistral models, with $\mathcal Y$ the vector space of the residual stream. We call these layers the \textit{probe layer} of the Qwen and Mistral models for reasons that will become clear. Due to the models' use of RMSNorm \cite{NEURIPS2019_1e8a1942}, the final activations have similar magnitude, and so we make the approximation throughout that they all lie on the $S^{n-1}$ manifold\footnote{Strictly speaking $S^{n-1}$ is the unit $n-1$ sphere, so we can either divide all $z_t$ by their constant norm (usually something like $\sqrt{n}$), or redefine the manifold as the $n-1$ sphere with radius equal to the $z_t$'s norm. The choice is arbitrary, but to lighten notation we will make it implicitly and keep speaking of $S^{n-1}$ as the manifold.}. Therefore $z_t$ can be restricted to $z_t\in S^{n-1}$. Finally, we define the vector $\mu\in T_{z_t}S^{n-1}$ to be in the tangent space of $S^{n-1}$ at $z_t$, and use the symbol when referring to any such tangent vector. 

Throughout we use chess as a controlled and inspect-able reasoning environment to analyse the model's behaviour. We do this by prompting it with Portable Game Notation (PGN) descriptions of a chess positions. We instruct it to think step-by-step and find the best move for the player whose turn it is. We collect a dataset of \textit{significant branch points} for chess, which are generation points where exchanging the top two tokens in otherwise greedy completions changes the final move the model recommends. At such points we also collect last hidden activations, the internal state at the probe layer, and an evaluation of the position (i.e.\ which side is winning) in centi-pawns using the Stockfish chess engine \cite{stockfish}. By brute forcing this problem, that is checking the output for each possible token swap, we produced $4,000$ such points. See figure \ref{fig:perturbation-analysis} for the exact prompt template and an example branch point.
 
As mentioned in the introduction, we wish to analyse the learned world model, and so we follow the interpretability ideas from OthelloGPT \cite{li2024emergentworldrepresentationsexploring, nanda2023emergentlinearrepresentationsworld, DBLP:journals/corr/abs-2310-07582} to train $737$ linear probes that correspond to binary classification tasks of labels like \texttt{mine\_knight\_f3} or \texttt{your\_pawn\_h4}. They classify $y\in \mathcal{Y}$, which is why we choose the name probe layer for layer $20$ and $30$ of the models. These particular layers were chosen as they gave us the best probe accuracy.

Using chess to investigate model state tracking has been explored in \cite{toshniwal2022chesstestbedlanguagemodel}. Emergent chess world models with linear probes were also explored in \cite{karvonen2024emergent}. Unlike the cited works, we do not train or fine tune the model in any way (extracting linear world representations from off-the-shelf models has also been explored in \cite{gurnee2024language}). A difficulty with this is that chess is extremely sparse: Taking a random walk through the tree of legal moves, a pawn, for instance, may only appear on h4 in a vanishing fraction of all positions. For this reason, we generate many legal random games and select a subset that balances the true/false labels for all the probes simultaneously.

Table \ref{tab:probe_metrics} shows an overview of the accuracies and F1 scores we managed to achieve on the balanced dataset for Qwen and table \ref{tab:probe_metrics_mistral} for Mistral. We note that the probe quality is significantly worse (but still much better than random) for the Mistral model, which will have implication for our analysis later.

\begin{table}[htbp]
\centering
\caption{Probe Performance Metrics (8 Representative Probes, Sorted by Accuracy) Qwen Model}
\label{tab:probe_metrics}
\input{probe_table}

\vspace{1em}
\caption{Probe Performance Metrics (4 Representative Probes, Sorted by Accuracy) Mistral Model}
\label{tab:probe_metrics_mistral}
\input{mistral_probe_table}
\end{table}
As a corollary, this shows that off-the-shelf instruct tuned language models implicitly learn a representation of chess. The linear probes have a form of $l(x) = w\cdot x + b$, where $w$ is a vector and $b$ is a scalar. We refer to $w$ as a \textit{world vector}. Prior work on analysing neural network representations is for instance \cite{zou2025representationengineeringtopdownapproach}. 

\section{Motivation}
Prior work \cite{bhargava2024whatsmagicwordcontrol,LuBM0S22} and our own investigation establishes that language models are significantly unstable to single token changes. Comparing the evaluation of the position after playing the move recommended by the \textit{greedy} and \textit{branch} completions (see section \ref{sec:nomenclature_and_notes} on how these are generated), the mean absolute centi-pawn change is $4.5\pm 1.5\ \text{log-centi-pawn}$ in log-space (a full Knight is approximately $\ln(300)\approx5.7$ log-centi-pawns), highlighting the significance of such changes.
\begin{figure}[t!]
    \input{example_completion_chess.tex}
    \caption{Impact of single-token perturbation at position 4 on reasoning and final move recommendation. Shows the completion before branching and then the difference between the continuations. In the difference, black, red, and green colours indicate coinciding, deleted, and new text in the perturbed completions.}
    \label{fig:perturbation-analysis}
\end{figure}

We note that the difference in word choice between the \textit{greedy} and \textit{branch} completions that causes a notably different move to be recommended seems incredibly minor (``analyze the \textit{position}'' vs ``analyze the \textit{step}'' in figure \ref{fig:perturbation-analysis}), and that the generation streams take many tokens before they truly begin to diverge in their output (see difference in figure \ref{fig:perturbation-analysis}).

While sensitivity to small perturbations is expected in auto-regressive systems (explored for instance here \cite{wang2024chain,Holtzman2020The}), we take the slightly more radical position that the phenomenon ought not to be dismissed entirely as exponential divergence due to compounding error (see prior work \cite{zhang2024how}). To that end we postulate that the geometry of blurry points is related to the internal state in a structured, semantically meaningful way; a new kind of relationship that is not captured by considering the exponential divergence alone.

\section{Developing the Geometry}
We postulate (and then empirically show) that treating
\begin{align}
    \label{eqn:three-blade}
    A(z_t) = 4z_t \wedge (p_1v_1) \wedge (p_2v_2)
\end{align}
as giving rise to a geometry that ``twists'' and turns the internal state $y$ in a precise manner leads to a measure of uncertainty with an intrinsic orientation. This is in contrast to scalar uncertainty measures like in \cite{kuhn2023semantic} or the entropy or variance, which do not possess a notion of orientation. Our key results show that this orientation of uncertainty \textit{couples} (i.e. absolute cosine similarity) to the world vectors we've extracted from the models' probe layers. While our construction is more akin to a gauge theory than general relativity, a geometric treatment of language models has also been made in \cite{disipio2025curvedspacetimetransformerarchitectures,Shao_2018_CVPR_Workshops}.

A point introduces a large error due to sampling when the angle between the predicted $z_t$ and the selected continuation embedding vector is large. We are not concerned with all generation points that have large sampling error: If $p_2$ is small despite $z_t$ and $v_1$ not being well aligned, then the error is high, but we have no choice; the second most likely token is too improbable to be sampled without causing artifacts in the model's output. On the other hand, when $v_1$ and $z_t$ are not aligned, and $p_2$ is reasonable, then it is possible to sample the second most likely token without breaking the auto-regressive process $F$. We are concerned with this second class of points, at which $z_t$ is between two possible continuations which may both reasonably be chosen.

To explain equation \ref{eqn:three-blade}, consider the parallelepiped (see figure \ref{fig:parallelepiped}) spanned by the vectors $z_t$, $(p_1v_1)$, and $(p_2v_2)$.
\begin{figure}[b]
    \centering
    \resizebox{0.40\textwidth}{!}{\input{assets/tikz/parallelapiped}}
    \caption{Parallelepiped spanned by $z_t$, $(p_1v_1)$, and $(p_2v_2)$ in three cases (left to right): Vectors are not aligned and probability mass is evenly distributed, probability mass is evenly distributed but vectors are aligned, vectors are not aligned but $p_2 \ll p_1$.}
    \label{fig:parallelepiped}
\end{figure}
The volume of the parallelepiped depends on the opening angle between all the vectors that span it and these vectors' lengths. If the vectors are aligned, the volume is small because the parallelepiped is shallow. If $(p_2v_2)$ is short because of low $p_2$ probability mass, then the volume is small because one of the object's faces becomes narrow. Its volume is only great if the vectors aren't aligned and the total probability mass is somewhat evenly distributed between $p_1$ and $p_2$. This is exactly the property of the blurry points we want to look at.

The three-blade of equation \ref{eqn:three-blade} contains this geometry algebraically. Particularly, the Frobenius norm of the wedge product gives the volume of the parallelepiped, such that
\begin{align*}
    ||A(z_t)||_F \propto \text{\small extent to which $z_t$ is a blurry point}.
\end{align*}

To develop equation \ref{eqn:three-blade} into defining a geometry that affects the internal state, we will upgrade it to a connection by contracting with the tangent vector $\mu$. In general, the connection is an $n \times n$ matrix that captures the infinitesimal change on $y$ due to the underlying geometry when moving (infinitesimally) in the direction $\mu$. In our case, the underlying geometry is defined by consideration of our blurring analogy as captured through equation \ref{eqn:three-blade}. Finally, contracting with $\mu$
\begin{align}
    \label{eqn:connection}
    A_\mu(z_t) &= 4p_1p_2( - (\mu \cdot v_1)(z_t \wedge v_2) + (\mu \cdot v_2)(z_t \wedge v_1)).
\end{align}
Where we've used that $\mu\cdot z_t = 0$, since $\mu$ is in the tangent space at $z_t$. This expression is a linear combination of two-blades. Using the isomorphism $\wedge^2(\mathbb R^n)\cong\text{Skew}_n(\mathbb R)$ (particularly $\phi(A\wedge B) = BA^T - AB^T$, where $\phi$ is an isomorphism), we notice that each two-blade is an element in the $\mathfrak{so}(n)$ Lie algebra, and so $A_\mu(z_t)$ is a $\mathfrak{so}(n)$ valued 1-form. Said another way, each two-blade (like $z_t\wedge v_2$) defines a parallelogram, the same way the three-blade gave a parallelepiped, with the matrix corresponding to, for instance, $z_t\wedge v_2$, being a generator\footnote{That is, the instantaneous velocity of the rotation. Since $SO(n)$ is a linear Lie group, if you apply the matrix exponential to this generator you are integrating following the velocities, and get a full $SO(n)$ rotation matrix corresponding to the generator back.} of rotations in the plane spanned by $z_t$ and $v_2$.

To formally treat equation \ref{eqn:connection} as a connection, we must attach a copy of $\mathcal Y$ to every point on the $S^{n-1}$ output manifold, with equation \ref{eqn:connection} acting as the connection between the independent vector spaces in this fibre bundle.

\subsection{Parallel Transport}
Equation \ref{eqn:connection} is the infinitesimal effect on $y$ due to the blurring geometry. The total change on $y$ due to blurring along a curve $\gamma$ on the $S^{n-1}$ manifold is given by the parallel transport. This is exactly how parallel transporting a state vector in the $U(1)$ fiber bundle of electromagnetism with the vector potential as the connection tells us how the electromagnetic field changes the internal phase of the quantum mechanical system.

The parallel transport operator along a curve defined by infinitesimal segments $\delta x_i$ is given by the repeated application of the matrix exponential of our connection
\begin{align}
    U = \lim_{N\to\infty}\prod^N_i \exp(-A_{\delta x_i}(z_i)),\label{eqn:U_as_a_product}
\end{align}
where each $z_i$ is given by the recursion $z_{i+1} = z_{i} + \delta x_i$. $A_\mu(z_t)$ is in principal smooth and differentiable for any $z_t$, as long as $A_\mu(z_t + \delta)$ does not change which two tokens are the top ones. This means (if one does not want to apply further tricks) the path $\gamma$ must be contained to a region around $z_t$ where the top two tokens are preserved.

Formally taking the limit in equation \ref{eqn:U_as_a_product} leads to the parallel transport operator for a curve $\gamma$ as the path ordered integral operator
\begin{align}
    U = P \exp\left(-\int_0^1 A_{\dot\gamma(s)}(\gamma(s)) ds\right). \label{eqn:U_as_integral}
\end{align}
With $P$ the path ordering symbol\footnote{Since each $A_{\dot\gamma(s)}(\gamma(s)) ds$ in equation \ref{eqn:U_as_integral} is a matrix we must be careful to maintain multiplication order. The path ordering symbol indicates that the exponentiated integral must resolve back to equation \ref{eqn:U_as_a_product} instead of something like $\exp(a + b + ...)$.} and where we explicitly parameterised the curve.

Given a path on the output manifold, we can now apply $U$ to a $y$ and see how the state vector changes due to the blurring geometry along the path.

\section{Probability Charge}
We would like to specifically highlight the $4p_1p_2$ product in our expression for $A_\mu(z_t)$. It mediates the coupling strength of our connection, and so, in analogy to the charge pre-factor in the connection of electromagnetism, we call it the \textit{probability charge}. 

In the same way that a chargeless particle does not experience the electromagnetic force due to that pre-factor being zero, an internal state does not experience the blurring geometry at a confident generation point when all the probability mass is concentrated in $p_1$, due to $p_2$ being zero. If we for a moment assume all the probability mass is concentrated between the top two tokens, then we may write $p_2\approx 1 - p_1$, and the probability charge becomes $4p_1(1 - p_1)$. This is an inverted parabola with a maximum of $1$ when $p_1 = 0.5$ --- a state experiences the blurring geometry maximally at maximum uncertainty.

\section{Holonomy}
We cannot take a state vector and parallel transport it from a $z_t$ to a $z_{t+1}$ in a meaningful way, as such a transport would require choosing a path which passes through many output states that the model never produced. We also cannot deform the output manifold to construct tunnels between $z_t$ and $z_{t+1}$ (see figure \ref{fig:wormholes}a), since we still need to choose a path through the ambient space that will ultimately traverse invalid intermediate states. In fact, with such a construction, the curvature experienced through the tunnel is exactly the accumulated curvature of the path we choose to construct it and so we wouldn't have resolved our path-dependence anyways.

\begin{figure}[t]
    \centering
    {\footnotesize a)}
    \resizebox{0.15\textwidth}{!}{\input{assets/tikz/manifold_tunnels}}
    {\footnotesize b)}
    \resizebox{0.15\textwidth}{!}{\input{assets/tikz/acc_holonomy}}
    \caption{(a) The manifold of the last hidden activations produced by the $F$ process. Each activation lies on the surface of an (ambient) sphere, and is connected via an instantaneous tunnel that it takes no time to travel through. (b) Two holonomy operators at two different points on the last hidden activation sphere. }
    \label{fig:wormholes}
\end{figure}

We can, however, parallel transport a state vector $y$ along a closed loop around a particular $z_t$, measuring only the local geometric effects. This is the holonomy operator, and it measures the extent to which closed loops on the manifold fail to close in the attached internal space due to the underlying geometry. We wish to use this to find physical effects due to the blurring geometry on the state vector $y$. One could construct such operators for each $z_t$ (see figure \ref{fig:wormholes}b) and apply them in time-ordered fashion to get the total holonomy due to a sequence up to time $t$
\begin{equation*}
    H_{\text{total}} = H_t H_{t-1}\dots H_{1}.
\end{equation*}

To compute $H$ for a particular $z_t$, we must choose a closed loop around $z_t$. Let us choose a plane at $z_t$ spanned by $u, v\in T_{z_t}S^{n-1}$, and construct the path in this plane. We wish to simply go around $z_t$ in a square with some small side length. However, choosing the obvious path (figure \ref{fig:clover}a) to do this introduces artifacts at order $\epsilon^3$ due to the local coordinate choice. A standard construction \cite{Gattringer2010_Sec9.9.2,alma990001919000206881} to resolve this is to compute an averaged clover like shown in figure \ref{fig:clover}b of four small squares with side length $\epsilon$. This yields $\mathcal{O}(\epsilon^4)$ accuracy.
\begin{figure}[t]
    \centering
    {\footnotesize a)} \resizebox{0.15\textwidth}{!}{\input{assets/tikz/holonomy_z_centred.tex}}
    {\footnotesize b)} \resizebox{0.14\textwidth}{!}{\input{assets/tikz/clover.tex}}
    \caption{(a) The path around $z_t$ we parallel transport along to find the holonomy. (b) A clover formed from four squares in the $u-v$ plane centred on a point $z_t$ to avoid artifacts due to the local coordinate choice.}
    \label{fig:clover}
\end{figure}
Notice that if $\epsilon$ is small enough, all points on the path defined by these squares are still approximately in the tangent space of $S^{n-1}$ at $z_t$\footnote{This is why the path for the squares in the clover does not need to bend along the sphere's surface.}. By this construction, the holonomy at a point $z_t$ is 
\begin{equation*}
    H_{z_t} = \frac{1}{4}(H^{(1)}_{z_t} + H^{(2)}_{z_t} + H^{(3)}_{z_t} +H^{(4)}_{z_t}),
\end{equation*}
where the upper index refers to the different squares in the clover. Each such square's holonomy is the path-ordered product of the parallel transport around its legs. To find each leg we must apply equation \ref{eqn:U_as_integral} four times with the path always being a straight line in the $u$ or $v$ direction. Since $\epsilon$ is small, we can Taylor expand the integrand in $-\int_0^1 A_{\dot\gamma(s)}(\gamma(s)){ds}$ in powers of $\epsilon$. As an example, for the top-right rectangle in the clover this leads to the legs
\begin{align*}
    L_1 &\approx \epsilon A_\mu(z),\quad L_2 \approx \epsilon A_\nu(z + \epsilon\mu),\\
    L_3 &\approx -\epsilon A_\mu(z + \epsilon\nu),\quad L_4 \approx -\epsilon A_\nu(z).
\end{align*}
where we've dropped all $\mathcal O(\epsilon^2)$ terms here. Using these, the holonomy operator for the top-right square in the clover then is
\begin{equation*}
    H^{(1)} = e^{-L_4}e^{-L_3}e^{-L_2}e^{-L_1}.
\end{equation*}
Now we can apply Baker-Campbell-Hausdorff\footnote{This formula gives how to combine products of exponentials into sums in the exponent for objects that don't commute. To first order it is $e^xe^y \approx e^{x + y + \nicefrac{1}{2}[x, y] + \dots}$} for each of the matrix exponentials, collect in powers of $\epsilon$, and sum the holonomy contributions from each square in the clover to arrive\footnote{The authors implemented this symbolically using SymPy \cite{10.7717/peerj-cs.103} to generate a JAX \cite{jax2018github} function for $H_{z_t}$} at the final holonomy operator at a point $z_t$
\begin{align} 
    \label{eqn:holonomy}
    H &= \exp(h)\nonumber\\
    h &\approx -\epsilon^2(\partial_\mu A_\nu(z) - \partial_\nu A_\mu(z) - [A_\nu(z), A_\mu(z)]) + \mathcal{O}(\epsilon^4),
\end{align}
where this being accurate to order $\epsilon^4$ despite only containing terms up to order $\epsilon^2$ is a consequence of the clover construction. Also note that we can recognise the term involving the partial derivatives and the matrix commutator\footnote{This is $[A, B] = AB - BA$} as the standard curvature tensor.

At each $z_t$ there are $n$ choose $2$ many planes tangent to the $S^{n-1}$ manifold. To avoid having to compute all of them, we only consider the plane with the largest magnitude contribution. We construct this by projecting $v_1$ onto the tangent space of $z_t$, forming $u$, and constructing $v$ as the projection of $v_2$ onto the tangent space at $z_t$ and orthogonalising to $u$.

The curvature that is captured as $h$ of our blurring geometry is a ``physically'' meaningful observable, which we will now test.

\section{Experimental Design}
In section \ref{sec:nomenclature_and_notes} we've already outlined our experimental context and have described how we have access to linear probes for the chess world model. We've also explained how we've generated the \textit{greedy} and \textit{branch} continuations.

At each \textit{branch point} we can compute a holonomy operator according to equation \ref{eqn:holonomy}. We also have access to two $y$ values: The first is the activation in the residual stream at the probe layer for the transition from the branch point to the next greedy token, and the second is the same but for the transition from the branch point to the branch token. We call these the \textit{greedy $y$} and \textit{branch $y$} respectively.

Since we have access to the true board state we prompted the model with, at all points we can choose to only look at the probes which are currently on the board or not. We call these the \textit{active} and \textit{bulk} probes.

Let us define the difference of applying $H$ and not as $q = Hy - y$ and note that $q = Hy - y = (H - \mathds 1)y \approx hy$, since $H=\exp(h)\approx 1 + h + ...$. We can recognize $h$ and consequently $H - \mathds 1$ (approximately\footnote{Since $h$ is $\mathcal{O}(\epsilon^2)$, keeping only the linear term in the Taylor series of the matrix exponential is $\mathcal{O}(\epsilon^2)$ accurate. Extending the Taylor series to the square order term would only introduce a $\mathcal{O}(\epsilon^4)$ correction.}) as the curvature of the blurring geometry.

$q$ is the vector one gets after applying the curvature due to blurring to $y$. One can simply treat it as an ``uncertainty vector'' if one so wishes, similar to how one can treat the entropy or variance as an ``uncertainty scalar,'' without invoking more of the underlying mathematics. Notice that in a well behaved theory we expect only the curvature to give us physical results --- the holonomy on its own is not an observable.

We call the absolute cosine similarity the \textit{coupling} between $q$ and the world vectors $w_i$.
\begin{figure}[t]
    \centering
    {\footnotesize a)}\includegraphics[width=0.44\textwidth]{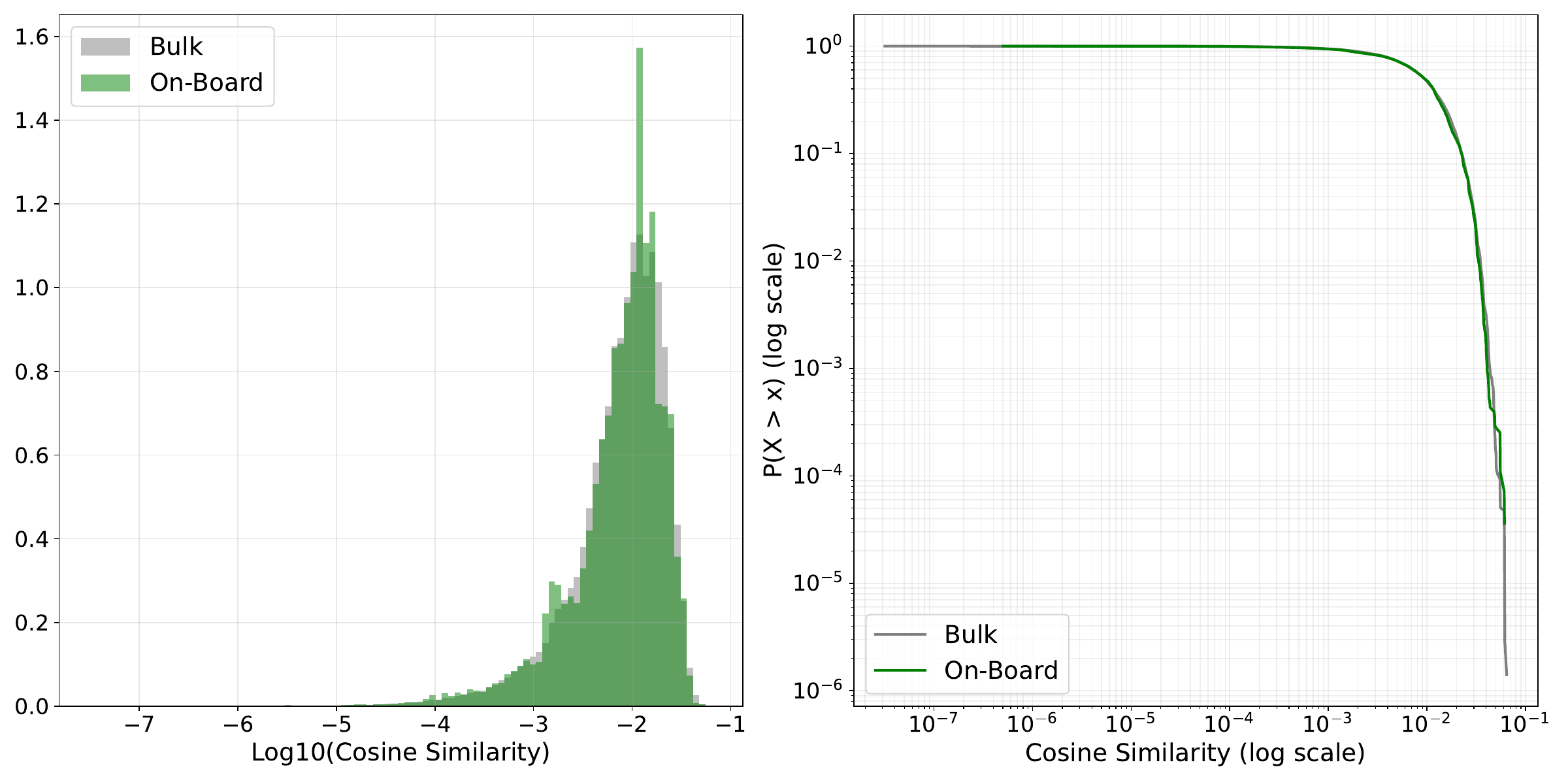}
    {\footnotesize b)}\includegraphics[width=0.22\textwidth]{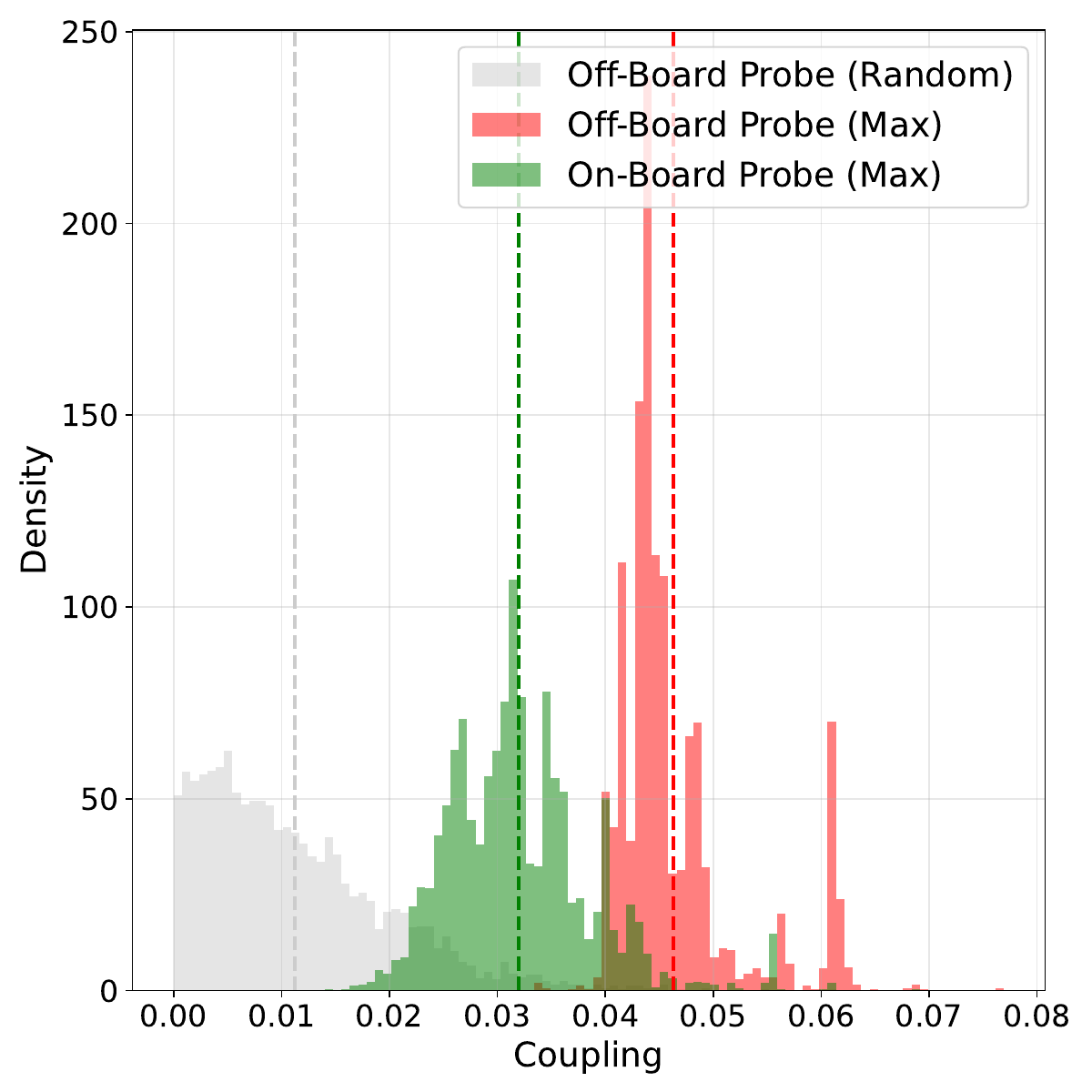}
    {\footnotesize c)}\includegraphics[width=0.22\textwidth]{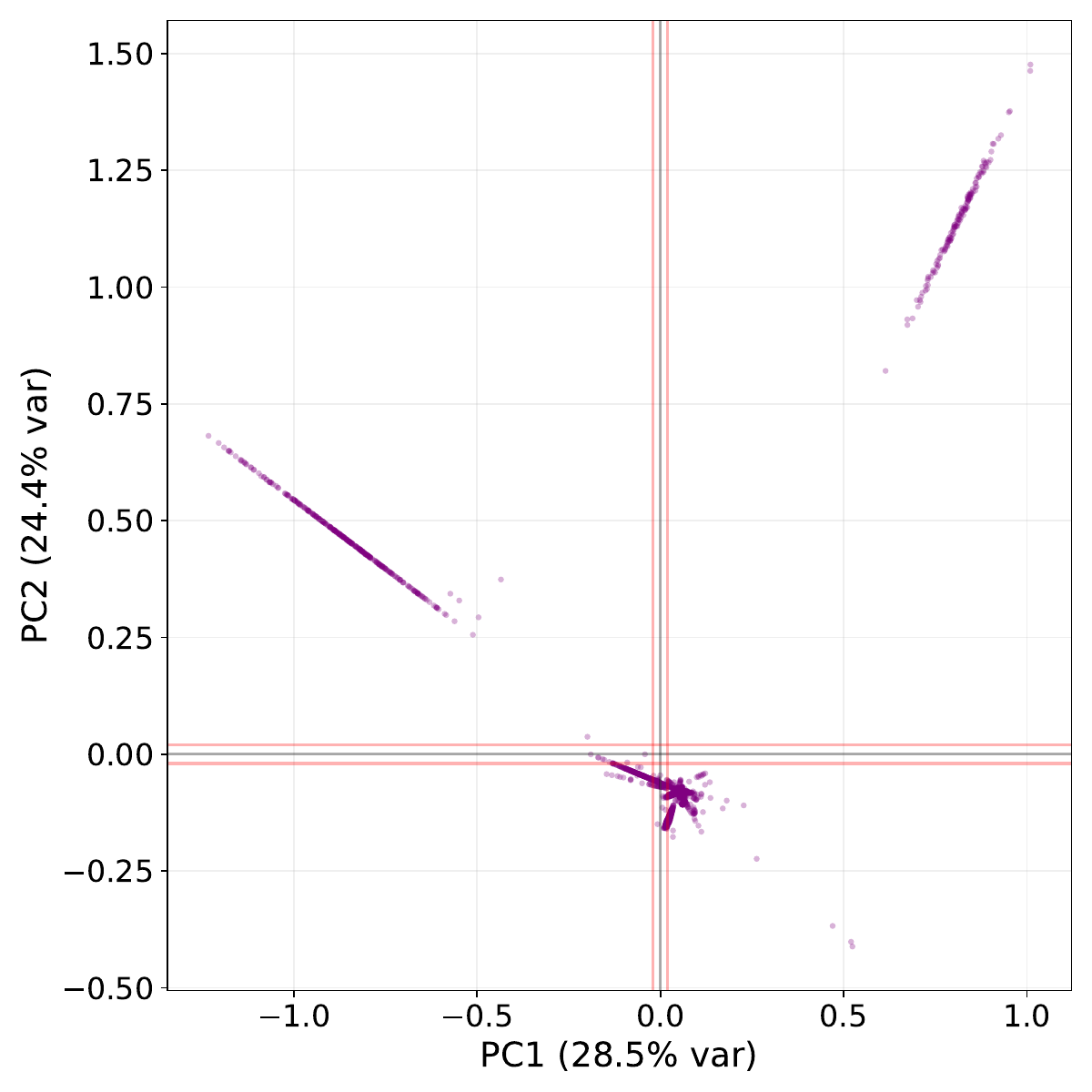}
    \caption{(a) Log space density and log-log survival plot of bulk and active distributions coupling strengths, showing they are inseparable. (b) Distribution of coupling strengths to random world vector from the bulk (gray), maximum probe from the active set (green), and maximum probe from the bulk set (red). (c) Two dimensional PCA scatter of maximal probe couplings.}
    \label{fig:probe_seperation_and_pca}
\end{figure}
We want to try to separate the probe distribution in some way. The first attempt is to simply consider the active and bulk set separately, but figure \ref{fig:probe_seperation_and_pca}a shows that they cannot be separated in this way. We can separate them however by considering the bulk and active set separately, and selecting the maximum coupling probe from these two sets (see figure \ref{fig:probe_seperation_and_pca}b). We will use the maximum coupling probe throughout.

\section{Results}
\subsection{Principal Component Analysis and Chess Board Files}
We ran two dimensional PCA on the population of $q$ vectors to see if there is semantic structure to them. This would indicate that the geometry of blurring derived purely from the token space geometry is semantically meaningful in an earlier layer, revealing a relationship between the blurring failure mode and the model's internal structure.
\begin{figure}[b]
    \centering
    {\footnotesize a)}\includegraphics[width=0.22\textwidth]{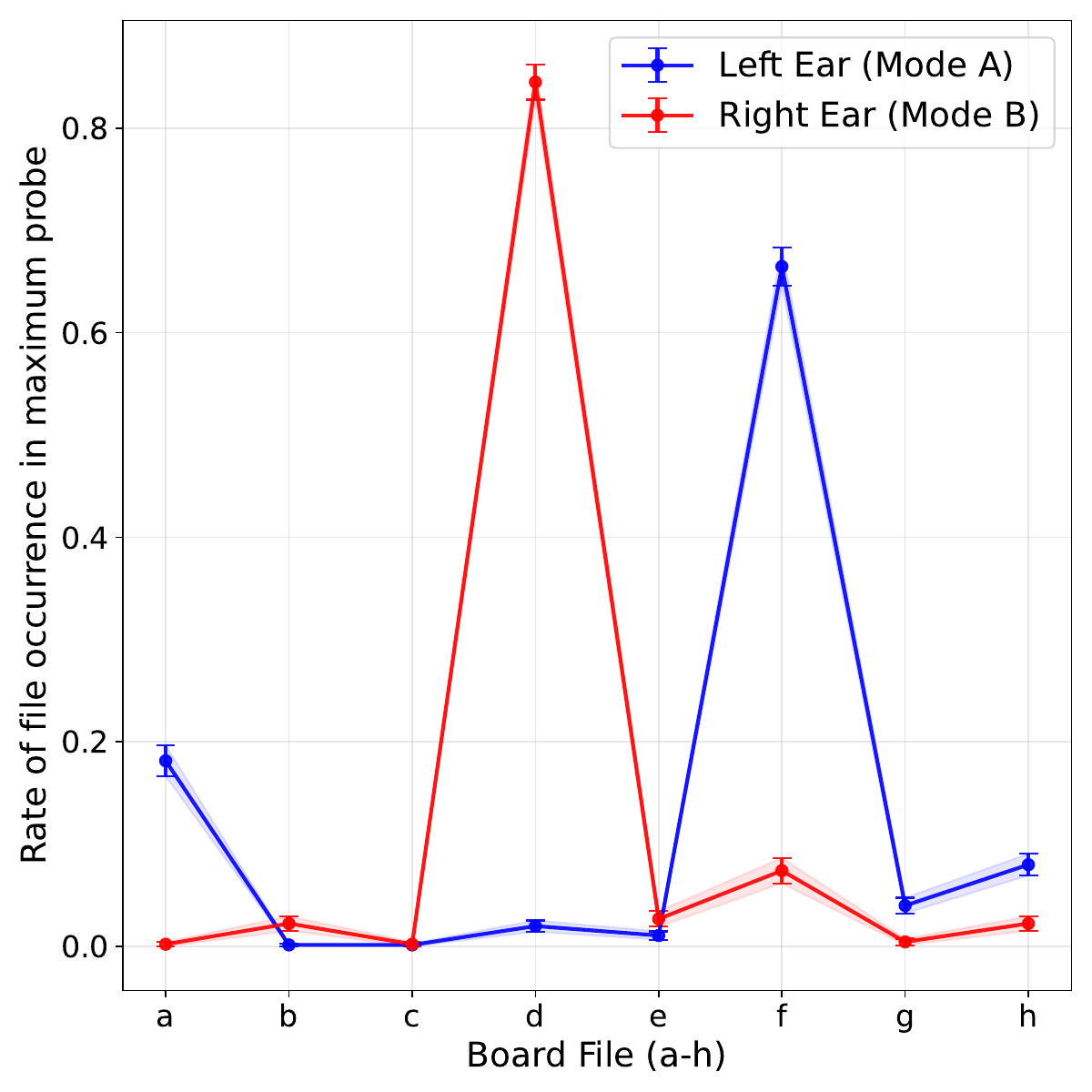}
    {\footnotesize b)}\includegraphics[width=0.22\textwidth]{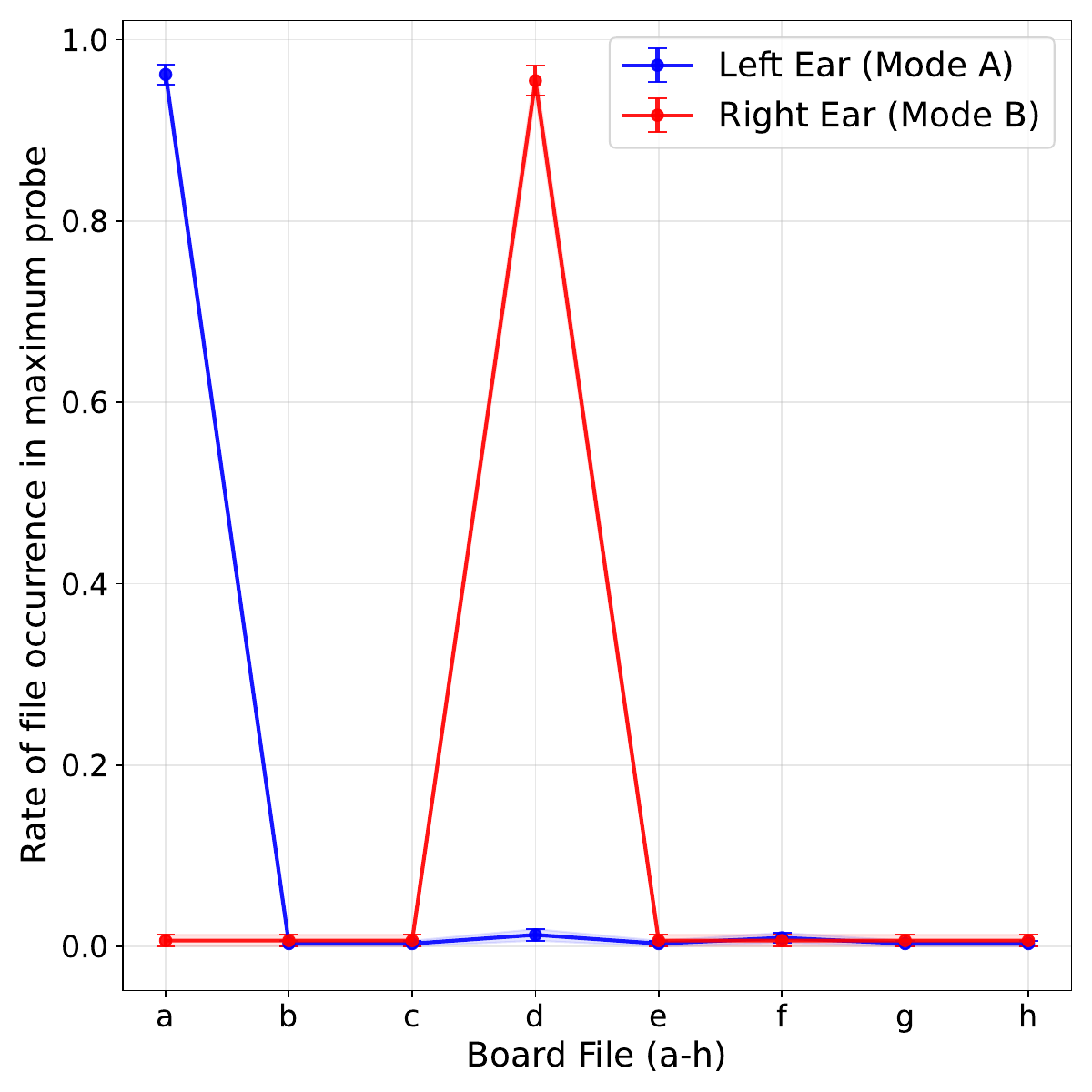}
    \caption{Distribution of chess file occurrences in maximum probes for left and right ear distributions for the greedy internal state (a) and branch internal state (b). Error is estimated using Bayesian methods by assuming a flat prior.}
    \label{fig:chess-files}
\end{figure}
Looking at figure \ref{fig:probe_seperation_and_pca}c for a scattering of the PCA results, we notice two ``ears'' of points in the top left and top right quadrants, away from the bulk in the centre. Selecting out the right and left ears and computing the coupling to the maximal probe, we can aggregate the result by which file (on the chess board a to h) the maximal probe feature belongs to. See figure \ref{fig:chess-files}a and b for the rate at which each file occurs for the greedy and branch $y$s. For both, we notice that they are separable distributions, with one ear peaking on the d file and the other on the f or a files. It appears that the rotation induced by $H$ is not random, but instead seems to separate king's \& queen's side and centre world vectors. Notice also that the scattering induced by $H$ for $y$s from the greedy continuation seems to mirror the semantic meaning of the one induced on the $y$s from the branch continuation. This appears to be a property of the holonomy operator directly, as it is the same in both cases, derived from the blurring geometry at the branch point, with only the $y$ it is being applied to changing.

\subsection{Three Dimensional PCA}
We can extend this approach straightforwardly to three dimensions.
\begin{figure}[t]
    {\footnotesize a)}\includegraphics[width=0.24\textwidth]{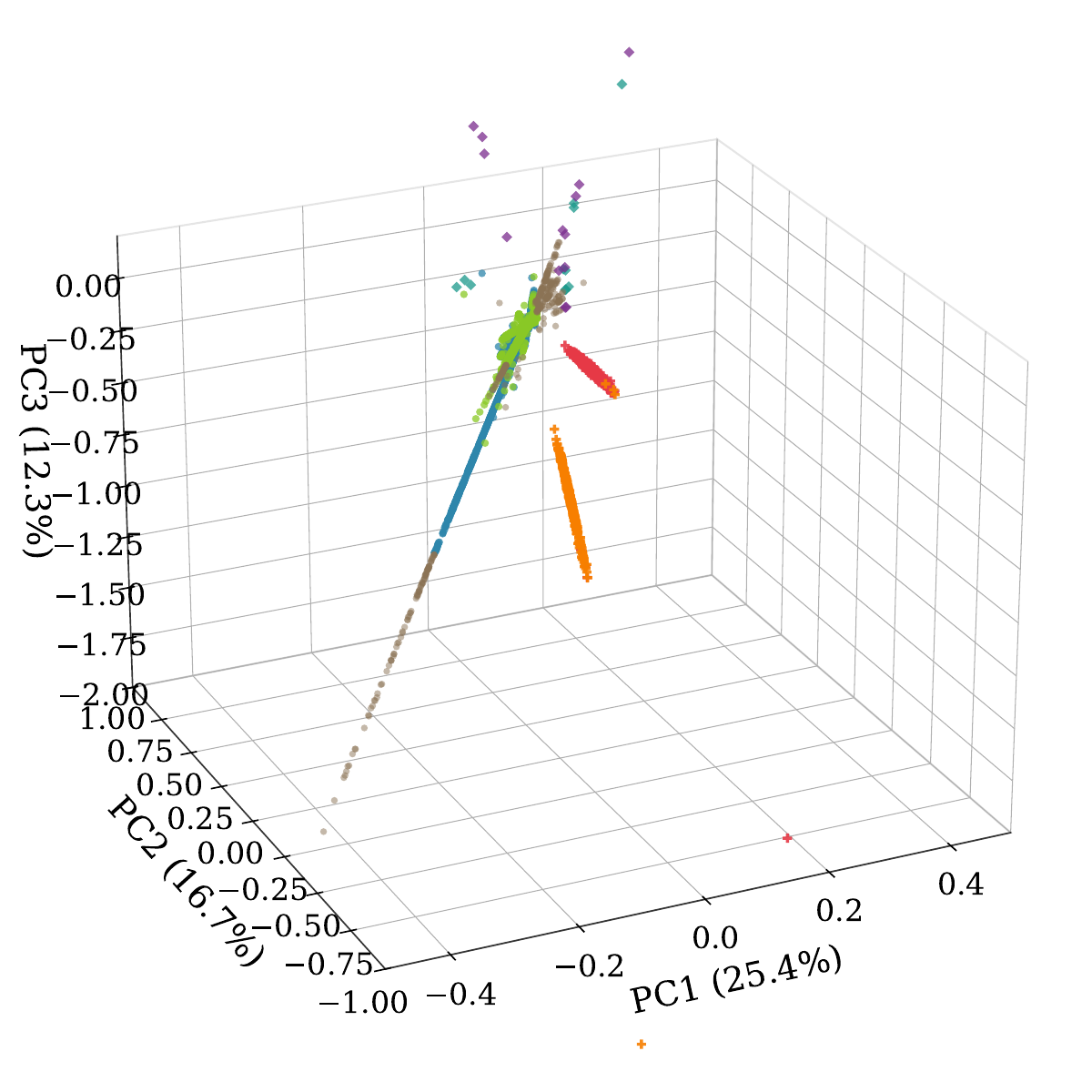}
    {\footnotesize b)}\includegraphics[width=0.20\textwidth]{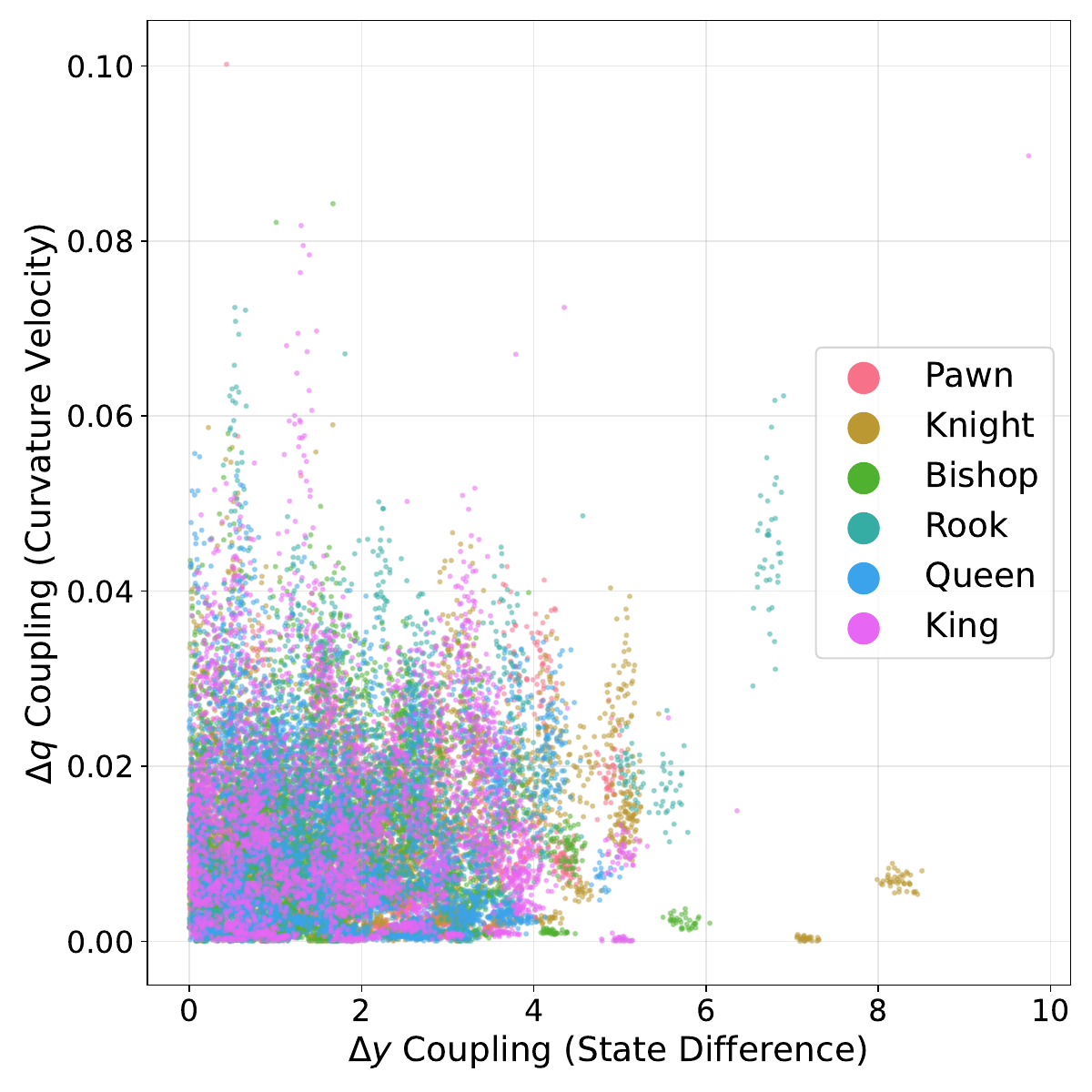}
     \caption{a) Scattering of coupling strengths as three dimensional PCA. Shades of blue and green indicate points belonging to left and right ear identified in two dimensional PCA. Red and yellow lines indicate greedy and branch points that form similar but separate lines. b) Coupling of $\Delta y$ to probes (by piece type) vs coupling of $\Delta q$ to same probes. Shows that the blurring geometry still twists the world model between the two possible continuations, even if the direct difference between the two branches' internal states is orthogonal to the world model.}
    \label{fig:pca-three-d_and_spectrum}
\end{figure}
This reveals a new set of interesting points (see figure \ref{fig:pca-three-d_and_spectrum}a) formed by two lines which are not classified by our existing ears structure, and separate themselves into one for the greedy and one for the branch continuations. That is, they are $q$ vectors where all the ones from a branch continuation lie in a line, alongside but entirely separate from a line formed by the greedy continuation $q$ vectors. We can use this as an opportunity to analyse the difference in the $q$ vectors between greedy and branch continuations.
\begin{figure}[b]
    \centering
    a)\includegraphics[width=0.22\textwidth]{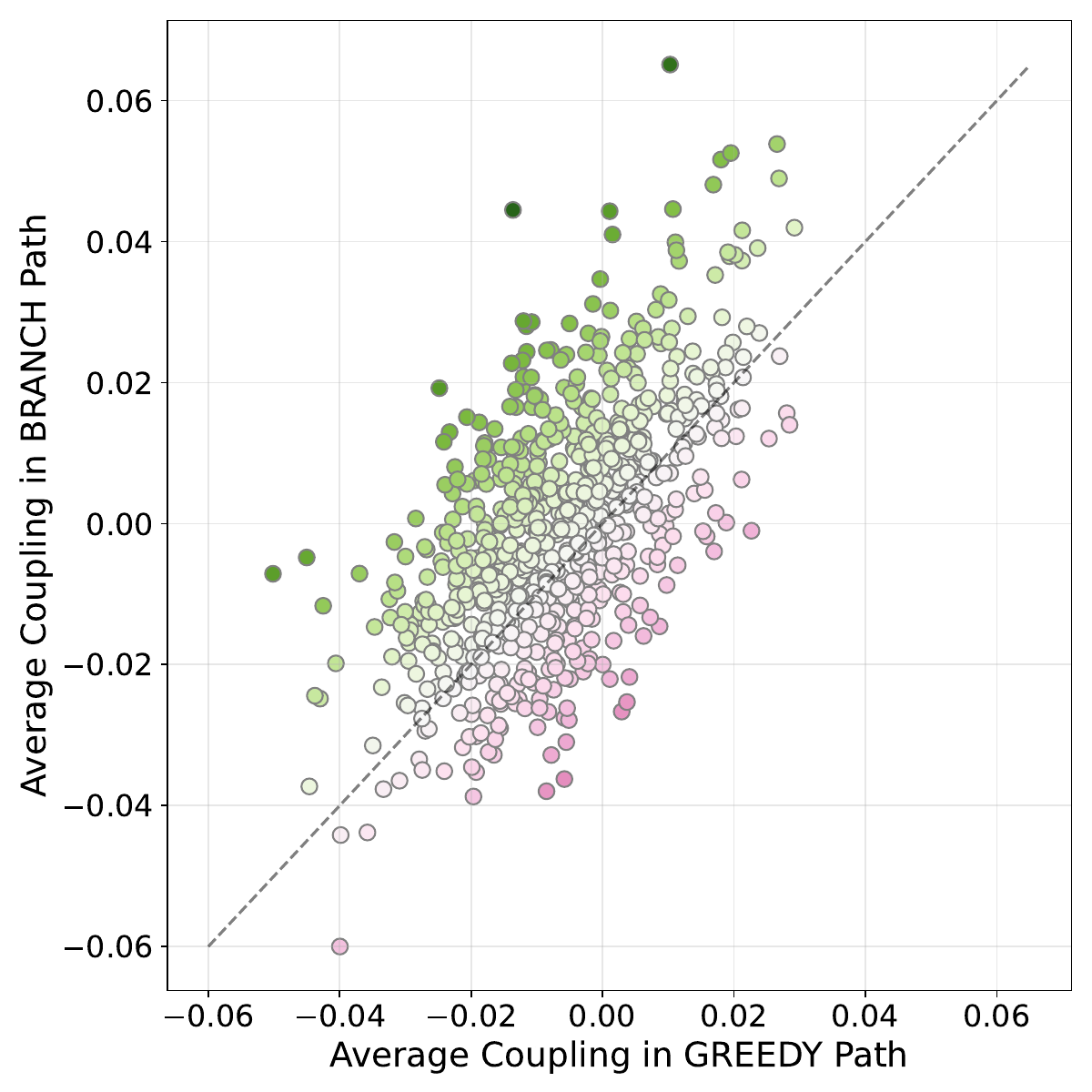}
    b)\includegraphics[width=0.22\textwidth]{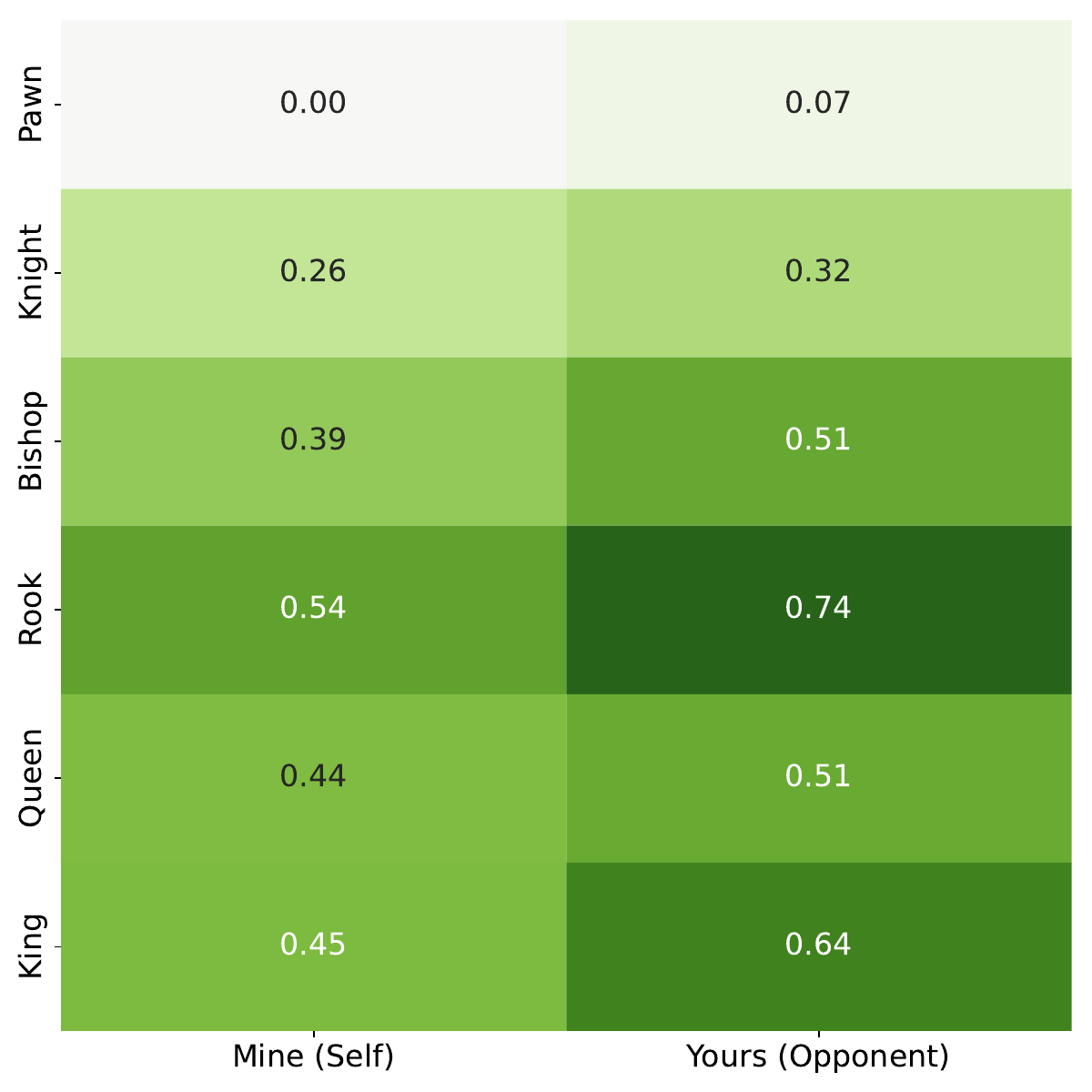}
    \caption{(a) Average probe coupling to $q$ vectors for the greedy and branch continuation plotted against each other. We note the systematically stronger coupling in the branch continuation. (b) Average coupling strength of $\Delta q$ vector to probes by piece type and opponent / own piece classification.}
    \label{fig:energy_and_semantic_shift}
\end{figure}

First, we compute the average coupling strength of the $q$s to all the probes for both the greedy and branch $y$s. We scatter these in figure \ref{fig:energy_and_semantic_shift}a. We notice a clear trend: The coupling in the branch continuation is systematically stronger than in the greedy completion. Again, we've applied the same $H$ computed with the same arguments to the different $y$s. It appears that the internal state of the branch path is systematically more entangled with the blurring curvature than that of the greedy branch. 

Let $q'$ be the $q$ vector of the branch path, while $q$ is this vector for the greedy path. We define $\Delta q = q' - q$, which measures the flow induced between the two internal states across the branch boundary. Figure \ref{fig:energy_and_semantic_shift}b shows the coupling strength of $\Delta q$ with the probes sorted by \textit{mine/yours} and the piece type. We notice that the coupling strength appears to follow the general importance of chess pieces (Spearman's\footnote{Computed by correlating coupling strength with standard piece value in pawns. We used $3.5$ pawns for the bishop and exclude the king.} $\rho=0.90$, $p=0.037$ for the \texttt{mine} pieces). With the coupling being almost entirely orthogonal to pawns, a little stronger for knights, then bishops, and finally strongest for the three most powerful pieces.

This means the difference in the rotation induced by $H$ for the two possible $y$s we apply it to (we stress that $H$ is the exact same matrix in both cases) couples to world vectors at a strength monotonically related to the relative chess piece importance.

\subsection{Internal Model Shift vs Holonomy Shift}
We can scatter the coupling of $\Delta q$ defined in the previous section vs the absolute shift in internal state $\Delta y = y' - y$.

Figure \ref{fig:pca-three-d_and_spectrum}b shows just this scatter. Notably, there is no clear trend line, and even for points on the left where $\Delta y$ is small, we see many points where $\Delta q$ is large. This means that despite the internal state shifting between the greedy and branch continuations in a manner that is orthogonal to the chess world model, our oriented measure of uncertainty does still rotate along the world model.

\subsection{Ablations}
To verify that the observed structure arises from the specific geometry of token-space blurring rather than arbitrary rotations, we applied a random $SO(n)$ matrix matching the Frobenius norm of the real $H$ operators to the same internal states. The resulting PCA shows no directional structure (see appendix figure \ref{fig:apx:holonomy_ablation}) and the two principal directions capture a lower fraction of the dataset variance ($53.9\%$ vs $34.8\%$), confirming that the semantic clustering is specific to our derived holonomy operator.

As a further ablation (see appendix figure \ref{fig:apx:holonomy_ablation}) we test applying an operator that rotates $z_t$ onto $v_2$ and one that rotates $z_t$ onto $v_1$. Similarly, we do not see a structure forming in the PCA and only $31.4\%$ of the total variance is captured in the two principal directions. It has been shown \cite{nostalgebraist2020logitlens,belrose2025elicitinglatentpredictionstransformers} that one can get semantically meaningful logit distributions by applying $V$ to intermediate layers. That a simple rotation like this does not give us an interesting PCA then is strong evidence that it is the tension between competing output vectors (i.e. blurring) and our particular operator form that is driving the structure we're observing.

Computing the coupling for $Hy$ instead of $q = Hy - y$ didn't lead to any semantically interesting results. As mentioned, the curvature of the field is the physical observable and so it is a sign that things are well behaved that we saw semantic coupling only for the $q$ population (Recall that $Hy - y \approx hy$, where we recognise $h$ as the curvature in equation \ref{eqn:holonomy}).

\subsection{Mistral}
As mentioned in section \ref{sec:nomenclature_and_notes}, we've reserved the Mistral model to analyse generalisability of our results.

See appendix figure \ref{fig:mistral_results} for a collection of the associated graphs. We note that, similar to the Qwen model, the probe couplings can be separated through the maximum \textit{active} and \textit{bulk} probes, and that the magnitude of the coupling strength is of similar order. Similarly again, running two dimensional PCA leads to clear separable lines. We notice that the magnitudes of the PCA1 and PCA2 axes are much smaller for the Mistral model than for the Qwen model. We believe that this is due to the poorer probe quality mentioned in section \ref{sec:nomenclature_and_notes}, causing the data to be more noisy. Because of this small magnitude, we've multiplied the coupling values by $400$ in the semantic shift and spectrum for the visualisation only. This doesn't affect Spearman's $\rho$ or the relative effect.

Notably, the three dimensional PCA also contains lines of separated \textit{greedy} and \textit{branch} points. Interestingly the flow for the Mistral model seems to be from the \textit{branch} to the \textit{greedy} states, opposite what we've observed for the Qwen model. Isolating these, we can run the same semantic shift analysis as before. We note that we see a similar relationship between piece importance and coupling strength ($\rho = -0.90$, $p=0.037$ for the \texttt{mine} pieces), except that the King appears to be an outlier (for Qwen and Mistral the piece was not included as it cannot be captured, and so no pawn value is commonly assigned to it). The spectrum result shows less structure than in the Qwen case (we didn't analyse this structure in either case), but does show the same interesting points where $\Delta y$ is small but $\Delta q$ is not.

Overall, the Mistral data confirms that our results are not a quirk of the Qwen model and that geometric treatment of blurring seems to generalise across model lineages. It is striking how similar the principle component structure looks in both cases.

\section{Conclusion}

We have introduced a novel geometric framework for understanding the relationship between token sampling ambiguity and language model internal states. Unlike scalar measures of uncertainty like entropy or variance, our framework provides an oriented quantity. We've shown that this orientation couples directly to the model's learned world model in ways related to the strategic nature of our reasoning task.

Our central contribution is that this measure is derived exclusively from the geometry of the token embeddings and yet successfully interrogates the complicated representation these general purpose LLMs learn. It implies that the indecision between possible next tokens is not a statistical fluke, while being a (of possibly many) precise --- structurally rich --- relationship between the output geometry and the model's internal representation.

We hope this geometric perspective contributes to a deeper understanding of how the discrete, combinatorial nature of language interfaces with the continuous geometry of neural networks, and believe this could lead to a more mechanistic understanding of the representations neural networks learn.

Our analysis is limited by being restricted to a single problem domain and only two general purpose models. Future research can attempt to relate the blurring geometry to reasoning tasks with harder to define strategic nature.

%% file: probe_table.tex
\small
\begin{tabular}{clcc}
\toprule
\textbf{Rank} & \textbf{Probe Label} & \textbf{Accuracy} & \textbf{F1 Score} \\
\midrule
1 & mine\_bishop\_on\_f1 & 0.825 & 0.826 \\
\multicolumn{4}{c}{$\dots$} \\
106 & mine\_bishop\_on\_c8 & 0.922 & 0.922 \\
211 & mine\_king\_on\_e6 & 0.956 & 0.957 \\
\multicolumn{4}{c}{$\dots$} \\
368 & mine\_knight\_on\_a2 & 0.966 & 0.966 \\
369 & mine\_knight\_on\_f2 & 0.966 & 0.967 \\
\multicolumn{4}{c}{$\dots$} \\
474 & yours\_rook\_on\_d2 & 0.969 & 0.967 \\
632 & mine\_rook\_on\_b3 & 0.978 & 0.977 \\
\multicolumn{4}{c}{$\dots$} \\
737 & mine\_king\_on\_a8 & 1.000 & 1.000 \\
\bottomrule
\end{tabular}

%% file: mistral_probe_table.tex
\small
\begin{tabular}{clcc}
\toprule
\textbf{Rank} & \textbf{Probe Label} & \textbf{Accuracy} & \textbf{F1 Score} \\
\midrule
1 & mine\_pawn\_on\_a4 & 0.760 & 0.763 \\
\multicolumn{4}{c}{$\dots$} \\
368 & mine\_knight\_on\_d2 & 0.887 & 0.888 \\
369 & mine\_knight\_on\_b8 & 0.887 & 0.885 \\
\multicolumn{4}{c}{$\dots$} \\
737 & yours\_king\_on\_a4 & 0.989 & 0.875 \\
\bottomrule
\end{tabular}

%% file: example_completion_chess.tex
\centering
\scriptsize  
\fbox{
\begin{minipage}{.43\textwidth}

\begin{center}
    \begin{minipage}[c]{0.5\textwidth}
        \centering
        \includegraphics[width=\linewidth]{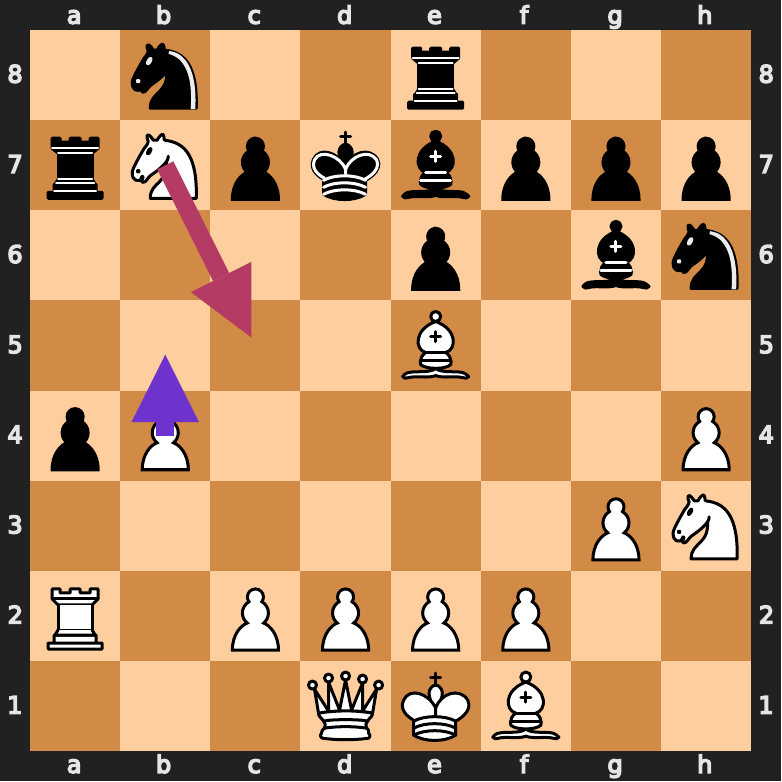}

    \end{minipage}
    \hfill 
    \begin{minipage}[c]{0.49\textwidth}
        \begin{adjustwidth}{0cm}{0cm}

            \textbf{Prompt}
            The following is the PGN of a chess game. You are playing as the white pieces, think step by step and find the best move. 1. h4 d6 2. Rh3 Be6 3. Re3 Bxa2 4. Rc3 a6 5. Rd3 b5 6. Rxd6 e6 7. Rxd8+ Kxd8 8. Nh3 Nh6 9. b4 Bd5 10. Ba3 Ra7 11. g3 Kd7 12. Nc3 Be7 13. Ra2 Be4 14. Nxb5 Bf5 15. Bb2 Re8 16. Nd6 a5 17. Nb7 Bg6 18. Be5 a4\\[2mm]
            \textbf{Completion before branch}\\
            Let's analyze the
        \end{adjustwidth}
    \end{minipage}
\end{center}
\textbf{Difference}\\
\textcolor{red!80!black}{\scriptsize [-  position step by -]} step to determine the best move for White.

\#\#\# Current Position:
- **White:** Kd1, Nb7, Nd6, Bb2, Be5, Ra2, g3, h4, b4, a2
- **Black:** Kd7, Nh6, Bf5, Be7, Be4, Ra7, Re8, a4, a5, b5, d6, e6

\#\#\# Key Points:
1. **Material:** White has a rook and two pawns more than Black.
2. **King Safety:** Both kings are relatively safe, but Black's king is more exposed.
3. **Piece Activity:** White's pieces are more active, especially the knights on b7 and d6.
4. **Pawn Structure:** White has a \textcolor{red!80!black}{\scriptsize [- passed -]}\textcolor{teal}{\textbf{[+ strong +]}} pawn on \textcolor{red!80!black}{\scriptsize [- the b-file, which is -]}\textcolor{teal}{\textbf{[+ b4, and Black has +]}} a \textcolor{red!80!black}{\scriptsize [- significant advantage. -]}\textcolor{teal}{\textbf{[+ weak pawn on a4. +]}}

\#\#\# Analysis:
- **White's \textcolor{red!80!black}{\scriptsize [- Passed Pawn:** The b4 -]}\textcolor{teal}{\textbf{[+ Advantage:** White has a significant material advantage and active pieces. - **Black's Weaknesses:** Black's king is not well-protected, and the +]}} pawn \textcolor{red!80!black}{\scriptsize [- is a passed pawn and -]}\textcolor{teal}{\textbf{[+ on a4 is weak. - **Potential Threats:** White +]}} can \textcolor{red!80!black}{\scriptsize [- be pushed -]}\textcolor{teal}{\textbf{[+ put pressure on Black's king and continue +]}} to \textcolor{red!80!black}{\scriptsize \textbf{-[diverged]-}}
\\[2mm]
\textbf{End of Completion}\\
    (...) The best move for White is **19. b5**. This move maximizes the advantage of the passed pawn and puts Black under significant pressure.\\[.5mm]
    (...) The best move for White is **19. Nc5**. This move maintains the initiative, puts pressure on Black's position, and continues to improve the activity of White's pieces.
\end{minipage}
}

%% file: assets/tikz/parallelapiped.tex
\begin{tikzpicture}[scale=1.2]
\usetikzlibrary{calc}
\begin{scope}[shift={(-4,0)}]

    \coordinate (O) at (0,0);
    \coordinate (zt) at (0,2.5);
    \coordinate (p1v1) at (1.8,1.2);
    \coordinate (p2v2) at (-1.5,1.5);
    
    \fill[blue!20,opacity=0.5] (O) -- (p1v1) -- ($(p1v1)+(zt)$) -- (zt) -- cycle;
    \fill[red!20,opacity=0.5] (O) -- (p2v2) -- ($(p2v2)+(zt)$) -- (zt) -- cycle;
    \fill[green!20,opacity=0.5] (O) -- (p1v1) -- ($(p1v1)+(p2v2)$) -- (p2v2) -- cycle;
    
    \fill[yellow!30,opacity=0.5] (zt) -- ($(zt)+(p1v1)$) -- ($(zt)+(p1v1)+(p2v2)$) -- ($(zt)+(p2v2)$) -- cycle;
    
    \fill[blue!30,opacity=0.5] (p2v2) -- ($(p2v2)+(p1v1)$) -- ($(p2v2)+(p1v1)+(zt)$) -- ($(p2v2)+(zt)$) -- cycle;
    \fill[red!30,opacity=0.5] (p1v1) -- ($(p1v1)+(p2v2)$) -- ($(p1v1)+(p2v2)+(zt)$) -- ($(p1v1)+(zt)$) -- cycle;
    
    \draw[thick] (O) -- (zt);
    \draw[thick] (O) -- (p1v1);
    \draw[thick] (O) -- (p2v2);
    \draw[thick] (zt) -- ($(zt)+(p1v1)$);
    \draw[thick] (zt) -- ($(zt)+(p2v2)$);
    \draw[thick] (p1v1) -- ($(p1v1)+(zt)$);
    \draw[thick] (p1v1) -- ($(p1v1)+(p2v2)$);
    \draw[thick] (p2v2) -- ($(p2v2)+(zt)$);
    \draw[thick] (p2v2) -- ($(p2v2)+(p1v1)$);
    \draw[thick] ($(zt)+(p1v1)$) -- ($(zt)+(p1v1)+(p2v2)$);
    \draw[thick] ($(zt)+(p2v2)$) -- ($(zt)+(p1v1)+(p2v2)$);
    \draw[thick] ($(p1v1)+(p2v2)$) -- ($(p1v1)+(p2v2)+(zt)$);
    
    \node[left] at ($(O)!0.5!(zt)$) {$z_t$};
    \node[below right] at ($(O)!0.6!(p1v1)$) {$(p_1v_1)$};
    \node[below left] at ($(O)!0.6!(p2v2)$) {$(p_2v_2)$};
    
    \fill (O) circle (2pt);
\end{scope}

\begin{scope}[shift={(0,0)}]

     \coordinate (O) at (0,0);
    \coordinate (zt) at (0,2.5);
    \coordinate (p1v1) at (0.6,1.0);
    \coordinate (p2v2) at (-0.2,1.6);
    
    \fill[blue!20,opacity=0.5] (O) -- (p1v1) -- ($(p1v1)+(zt)$) -- (zt) -- cycle;
    \fill[red!20,opacity=0.5] (O) -- (p2v2) -- ($(p2v2)+(zt)$) -- (zt) -- cycle;
    \fill[green!20,opacity=0.5] (O) -- (p1v1) -- ($(p1v1)+(p2v2)$) -- (p2v2) -- cycle;
    
    \fill[yellow!30,opacity=0.5] (zt) -- ($(zt)+(p1v1)$) -- ($(zt)+(p1v1)+(p2v2)$) -- ($(zt)+(p2v2)$) -- cycle;
    
    \fill[blue!30,opacity=0.5] (p2v2) -- ($(p2v2)+(p1v1)$) -- ($(p2v2)+(p1v1)+(zt)$) -- ($(p2v2)+(zt)$) -- cycle;
    \fill[red!30,opacity=0.5] (p1v1) -- ($(p1v1)+(p2v2)$) -- ($(p1v1)+(p2v2)+(zt)$) -- ($(p1v1)+(zt)$) -- cycle;
    
    \draw[thick] (O) -- (zt);
    \draw[thick] (O) -- (p1v1);
    \draw[thick] (O) -- (p2v2);
    \draw[thick] (zt) -- ($(zt)+(p1v1)$);
    \draw[thick] (zt) -- ($(zt)+(p2v2)$);
    \draw[thick] (p1v1) -- ($(p1v1)+(zt)$);
    \draw[thick] (p1v1) -- ($(p1v1)+(p2v2)$);
    \draw[thick] (p2v2) -- ($(p2v2)+(zt)$);
    \draw[thick] (p2v2) -- ($(p2v2)+(p1v1)$);
    \draw[thick] ($(zt)+(p1v1)$) -- ($(zt)+(p1v1)+(p2v2)$);
    \draw[thick] ($(zt)+(p2v2)$) -- ($(zt)+(p1v1)+(p2v2)$);
    \draw[thick] ($(p1v1)+(p2v2)$) -- ($(p1v1)+(p2v2)+(zt)$);
    
    \node[left] at ($(O)!0.5!(zt)$) {$z_t$};
    \node[right] at ($(O)!0.7!(p1v1)$) {$(p_1v_1)$};
    \node[left] at ($(O)!0.7!(p2v2)$) {$(p_2v_2)$};
    
    \fill (O) circle (2pt);
\end{scope}

\begin{scope}[shift={(4,0)}]
    
    \coordinate (O) at (0,0);
    \coordinate (zt) at (0,2.5);
    \coordinate (p1v1) at (1.8,1.2);
    \coordinate (p2v2) at (-0.3,0.3); 

    \fill[blue!20,opacity=0.5] (O) -- (p1v1) -- ($(p1v1)+(zt)$) -- (zt) -- cycle;
    \fill[red!20,opacity=0.5] (O) -- (p2v2) -- ($(p2v2)+(zt)$) -- (zt) -- cycle;
    \fill[green!20,opacity=0.5] (O) -- (p1v1) -- ($(p1v1)+(p2v2)$) -- (p2v2) -- cycle;
    
    \fill[yellow!30,opacity=0.5] (zt) -- ($(zt)+(p1v1)$) -- ($(zt)+(p1v1)+(p2v2)$) -- ($(zt)+(p2v2)$) -- cycle;
    
    \fill[blue!30,opacity=0.5] (p2v2) -- ($(p2v2)+(p1v1)$) -- ($(p2v2)+(p1v1)+(zt)$) -- ($(p2v2)+(zt)$) -- cycle;
    \fill[red!30,opacity=0.5] (p1v1) -- ($(p1v1)+(p2v2)$) -- ($(p1v1)+(p2v2)+(zt)$) -- ($(p1v1)+(zt)$) -- cycle;
    
    \draw[thick] (O) -- (zt);
    \draw[thick] (O) -- (p1v1);
    \draw[thick] (O) -- (p2v2);
    \draw[thick] (zt) -- ($(zt)+(p1v1)$);
    \draw[thick] (zt) -- ($(zt)+(p2v2)$);
    \draw[thick] (p1v1) -- ($(p1v1)+(zt)$);
    \draw[thick] (p1v1) -- ($(p1v1)+(p2v2)$);
    \draw[thick] (p2v2) -- ($(p2v2)+(zt)$);
    \draw[thick] (p2v2) -- ($(p2v2)+(p1v1)$);
    \draw[thick] ($(zt)+(p1v1)$) -- ($(zt)+(p1v1)+(p2v2)$);
    \draw[thick] ($(zt)+(p2v2)$) -- ($(zt)+(p1v1)+(p2v2)$);
    \draw[thick] ($(p1v1)+(p2v2)$) -- ($(p1v1)+(p2v2)+(zt)$);
    
    \node[left] at ($(O)!0.5!(zt)$) {$z_t$};
    \node[below right] at ($(O)!0.6!(p1v1)$) {$(p_1v_1)$};
    \node[below left] at ($(O)!0.8!(p2v2)$) {$(p_2v_2)$};
    
    \fill (O) circle (2pt);
\end{scope}

\end{tikzpicture}

%% file: assets/tikz/manifold_tunnels.tex
\tdplotsetmaincoords{70}{110}

\begin{tikzpicture}[tdplot_main_coords, scale=3]

    \def\R{1}

    \draw[thick, tdplot_screen_coords] (0,0) circle (\R);

    
    \draw[gray] (\R,0,0) arc (0:-180:\R);
    \draw[dashed, gray] (\R,0,0) arc (0:180:\R);

    \tdplotsetthetaplanecoords{70}
    \draw[dashed, gray, tdplot_rotated_coords] (1,0) arc (0:180:1);
    \draw[gray, tdplot_rotated_coords] (1,0) arc (0:-180:1);

    \tdplotsetcoord{Pzt}{1}{55}{150} 
    
    \tdplotsetcoord{Pzt1}{1}{60}{20}
    
    \tdplotsetcoord{Pzt2}{1}{140}{10}

    
    \draw[thick, double, double distance=2pt] (Pzt) to[bend left=25] (Pzt1);
    
    \draw[thick, double, double distance=2pt] (Pzt1) to[bend left=20] (Pzt2);

    \fill[black] (Pzt) circle (1.5pt) node[left, xshift=-1mm, yshift=0] {\Large $z_t$};
    \fill[black] (Pzt1) circle (1.5pt) node[above right, xshift=-5mm, yshift=1mm] {\Large $z_{t+1}$};
    \fill[black] (Pzt2) circle (1.5pt) node[right, xshift=1mm, yshift=1mm] {\Large $z_{t+2}$};


\end{tikzpicture}

%% file: assets/tikz/acc_holonomy.tex
\tdplotsetmaincoords{70}{110}

\begin{tikzpicture}[tdplot_main_coords, scale=3]
    \def\R{1} 

    \draw[thin, tdplot_screen_coords] (0,0) circle (\R);

    \draw[gray, thin, dashed] (\R,0,0) arc (0:180:\R);
    \draw[gray, thin] (\R,0,0) arc (0:-180:\R);
    
    \tdplotsetthetaplanecoords{0}
    \draw[gray, thin, dashed, tdplot_rotated_coords] (0,1,0) arc (90:270:1);
    \draw[gray, thin, tdplot_rotated_coords] (0,1,0) arc (90:-90:1);

    \def\thetaOne{60}
    \def\phiOne{180}
    \tdplotsetcoord{P1}{\R}{\thetaOne}{\phiOne}

    \def\thetaTwo{120}
    \def\phiTwo{-10}
    \tdplotsetcoord{P2}{\R}{\thetaTwo}{\phiTwo}

    \def\d{0.25} 
    \definecolor{pencilblue}{RGB}{3,28,252}

    \tdplotsetrotatedcoords{\phiOne}{\thetaOne}{45}
    \begin{scope}[tdplot_rotated_coords, shift={(0,0,\R)}]
        \draw[thick, pencilblue] 
            (-\d, -\d, 0) -- (\d, -\d, 0) -- (\d, \d, 0) -- (-\d, \d, 0) -- cycle;
        \draw[thick, pencilblue, ->, >={Stealth[scale=1.5]}] 
            (-\d, -\d, 0) -- (\d, -\d, 0);
    \end{scope}

    \tdplotsetrotatedcoords{\phiTwo}{\thetaTwo}{30}
    \begin{scope}[tdplot_rotated_coords, shift={(0,0,\R)}]
        \draw[thick, pencilblue] 
            (-\d, -\d, 0) -- (\d, -\d, 0) -- (\d, \d, 0) -- (-\d, \d, 0) -- cycle;
        \draw[thick, pencilblue, ->, >={Stealth[scale=1.5]}] 
            (-\d, -\d, 0) -- (\d, -\d, 0);
    \end{scope}

    \fill[black] (P1) circle (1.5pt);
    \fill[black] (P2) circle (1.5pt);

    \node[anchor=south, xshift=1mm, yshift=0.5mm] at (P1) {\large $z_t$};
    \node[anchor=north west,  xshift=0, yshift=0] at (P2) {\large $z_{t+1}$};

\end{tikzpicture}

%% file: assets/tikz/holonomy_z_centred.tex
\begin{tikzpicture}[
    scale=2,
    thick,
    midarrow/.style={
        decoration={
            markings,
            mark=at position 0.55 with {\arrow{>}}
        },
        postaction={decorate}
    }
]

    \coordinate (BL) at (-1, -1); 
    \coordinate (TL) at (-1, 1);  
    \coordinate (TR) at (1, 1);   
    \coordinate (BR) at (1, -1);  
    \coordinate (Center) at (0, 0);

    \draw[midarrow] (BL) -- (TL);
    \draw[midarrow] (TL) -- (TR);
    \draw[midarrow] (TR) -- (BR);
    \draw[midarrow] (BR) -- (BL);

    \fill (Center) circle (2pt) node[below right, yshift=-2pt] {\Large $z_t$};

    \node at (BR) [below right] {\Large $\gamma$};

\end{tikzpicture}

%% file: assets/tikz/clover.tex
\begin{tikzpicture}[
    scale=3,
    thick,
    arrow ccw/.style={
        decoration={
            markings,
            mark=at position 0.55 with {
                \arrow[scale=2]{Stealth}
            }
        },
        postaction={decorate}
    },
    arrow cw/.style={
        decoration={
            markings,
            mark=at position 0.55 with {
                \arrow[scale=2]{Stealth}
            }
        },
        postaction={decorate}
    }
]

    \coordinate (Center) at (0,0);
    \def\sz{1} 

    \draw[arrow ccw] (-\sz, 0) -- (0, 0);     
    \draw[arrow ccw] (0, 0) -- (0, \sz);      
    \draw[arrow ccw] (0, \sz) -- (-\sz, \sz); 
    \draw[arrow ccw] (-\sz, \sz) -- (-\sz, 0);

    \draw[arrow cw] (\sz, 0) -- (0, 0);      
    \draw[arrow cw] (0, 0) -- (0, \sz);      
    \draw[arrow cw] (0, \sz) -- (\sz, \sz);  
    \draw[arrow cw] (\sz, \sz) -- (\sz, 0);  

    \draw[arrow ccw] (\sz, 0) -- (0, 0);       
    \draw[arrow ccw] (0, 0) -- (0, -\sz);      
    \draw[arrow ccw] (0, -\sz) -- (\sz, -\sz); 
    \draw[arrow ccw] (\sz, -\sz) -- (\sz, 0);  

    \draw[arrow cw] (-\sz, 0) -- (0, 0);       
    \draw[arrow cw] (0, 0) -- (0, -\sz);       
    \draw[arrow cw] (0, -\sz) -- (-\sz, -\sz); 
    \draw[arrow cw] (-\sz, -\sz) -- (-\sz, 0); 

    \fill (Center) circle (1.5pt) node[below right, xshift=2pt] {\Large $z_t$};

\end{tikzpicture}

%% file: main.bib
@misc{nanda2023emergentlinearrepresentationsworld,
      title={Emergent Linear Representations in World Models of Self-Supervised Sequence Models}, 
      author={Neel Nanda and Andrew Lee and Martin Wattenberg},
      year={2023},
      eprint={2309.00941},
      archivePrefix={arXiv},
      primaryClass={cs.LG},
      url={https://arxiv.org/abs/2309.00941}, 
}

@misc{li2024emergentworldrepresentationsexploring,
      title={Emergent World Representations: Exploring a Sequence Model Trained on a Synthetic Task}, 
      author={Kenneth Li and Aspen K. Hopkins and David Bau and Fernanda Viégas and Hanspeter Pfister and Martin Wattenberg},
      year={2024},
      eprint={2210.13382},
      archivePrefix={arXiv},
      primaryClass={cs.LG},
      url={https://arxiv.org/abs/2210.13382}, 
}

@inbook{Gattringer2010_Sec9.9.2,
      author    = {Gattringer, Christof and Lang, Christian B.},
      title     = {Quantum Chromodynamics on the Lattice},
      subtitle  = {An Introductory Presentation},
      booktitle = {Quantum Chromodynamics on the Lattice: An Introductory Presentation},
      year      = {2010},
      publisher = {Springer Berlin Heidelberg},
      address   = {Berlin, Heidelberg},
      series    = {Lecture Notes in Physics},
      volume    = {788},
      doi       = {10.1007/978-3-642-01850-3},
      isbn      = {978-3-642-01849-7},
      chapter   = {9},
      note      = {Section 9.9.2},
}

@book{alma990001919000206881,
      language = {eng},
      lccn = {94003438},
      publisher = {World Scientific},
      series = {K \& E series on knots and everything ; v.4},
      title = {Gauge fields, knots and gravity },
      address = {Singapore ;},
      booktitle = {Gauge fields, knots and gravity},
      isbn = {9810217293},
      author = {Baez, John C. and Muniain, Javier P.},
      year = {1994},
      note = {Exercise 95}
}

@article{radford2018improving,
  title={Improving Language Understanding by Generative Pre-Training},
  author={Radford, Alec and Narasimhan, Karthik and Salimans, Tim and Sutskever, Ilya},
  journal={OpenAI Technical Report},
  year={2018},
  url={https://cdn.openai.com/research-covers/language-unsupervised/language_understanding_paper.pdf}
}

@inproceedings{NEURIPS2019_1e8a1942,
 author = {Zhang, Biao and Sennrich, Rico},
 booktitle = {Advances in Neural Information Processing Systems},
 editor = {H. Wallach and H. Larochelle and A. Beygelzimer and F. d\textquotesingle Alch\'{e}-Buc and E. Fox and R. Garnett},
 pages = {},
 publisher = {Curran Associates, Inc.},
 title = {Root Mean Square Layer Normalization},
 url = {https://proceedings.neurips.cc/paper_files/paper/2019/file/1e8a19426224ca89e83cef47f1e7f53b-Paper.pdf},
 volume = {32},
 year = {2019}
}

@misc{qwen2.5,
    title = {Qwen2.5: A Party of Foundation Models},
    url = {https://qwenlm.github.io/blog/qwen2.5/},
    author = {{Qwen Team}},
    month = {September},
    year = {2024}
}

@misc{disipio2025curvedspacetimetransformerarchitectures,
      title={The Curved Spacetime of Transformer Architectures}, 
      author={Riccardo Di Sipio and Jairo Diaz-Rodriguez and Luis Serrano},
      year={2025},
      eprint={2511.03060},
      archivePrefix={arXiv},
      primaryClass={cs.LG},
      url={https://arxiv.org/abs/2511.03060}, 
}

@misc{toshniwal2022chesstestbedlanguagemodel,
      title={Chess as a Testbed for Language Model State Tracking}, 
      author={Shubham Toshniwal and Sam Wiseman and Karen Livescu and Kevin Gimpel},
      year={2022},
      eprint={2102.13249},
      archivePrefix={arXiv},
      primaryClass={cs.CL},
      url={https://arxiv.org/abs/2102.13249}, 
}

@inproceedings{
karvonen2024emergent,
title={Emergent World Models and Latent Variable Estimation in Chess-Playing Language Models},
author={Adam Karvonen},
booktitle={First Conference on Language Modeling},
year={2024},
url={https://openreview.net/forum?id=PPTrmvEnpW}
}

@misc{bhargava2024whatsmagicwordcontrol,
      title={What's the Magic Word? A Control Theory of LLM Prompting}, 
      author={Aman Bhargava and Cameron Witkowski and Shi-Zhuo Looi and Matt Thomson},
      year={2024},
      eprint={2310.04444},
      archivePrefix={arXiv},
      primaryClass={cs.CL},
      url={https://arxiv.org/abs/2310.04444}, 
}

@software{stockfish,
  author = {{The Stockfish developers}},
  title = {Stockfish},
  url = {https://stockfishchess.org/},
  license = {GPL-3.0}
}

@misc{mistralsmall31,
  title = {Mistral Small 3.1},
  author = {{Mistral AI Team}},
  year = {2025},
  month = mar,
  howpublished = {Mistral AI News},
  url = {https://mistral.ai/news/mistral-small-3-1/},
  note = {Model ID: mistralai/Mistral-Small-3.1-24B-Instruct-2503}
}

@article{wang2024chain,
  title={Chain-of-thought reasoning without prompting},
  author={Wang, Xuezhi and Zhou, Denny},
  journal={Advances in Neural Information Processing Systems},
  volume={37},
  pages={66383--66409},
  year={2024}
}

@misc{nostalgebraist2020logitlens,
  title = {interpreting {GPT}: the logit lens},
  author = {nostalgebraist},
  year = {2020},
  howpublished = {\url{https://www.lesswrong.com/posts/AcKRB8wDpdaN6v6ru/interpreting-gpt-the-logit-lens}},
  note = {Accessed: 2025-01-28}
}

@software{jax2018github,
  author = {James Bradbury and Roy Frostig and Peter Hawkins and Matthew James Johnson and Chris Leary and Dougal Maclaurin and George Necula and Adam Paszke and Jake Vander{P}las and Skye Wanderman-{M}ilne and Qiao Zhang},
  title = {{JAX}: composable transformations of {P}ython+{N}um{P}y programs},
  url = {http://github.com/jax-ml/jax},
  version = {0.3.13},
  year = {2018},
}

@article{10.7717/peerj-cs.103,
     title = {SymPy: symbolic computing in Python},
     author = {Meurer, Aaron and Smith, Christopher P. and Paprocki, Mateusz and \v{C}ert\'{i}k, Ond\v{r}ej and Kirpichev, Sergey B. and Rocklin, Matthew and Kumar, AMiT and Ivanov, Sergiu and Moore, Jason K. and Singh, Sartaj and Rathnayake, Thilina and Vig, Sean and Granger, Brian E. and Muller, Richard P. and Bonazzi, Francesco and Gupta, Harsh and Vats, Shivam and Johansson, Fredrik and Pedregosa, Fabian and Curry, Matthew J. and Terrel, Andy R. and Rou\v{c}ka, \v{S}t\v{e}p\'{a}n and Saboo, Ashutosh and Fernando, Isuru and Kulal, Sumith and Cimrman, Robert and Scopatz, Anthony},
     year = 2017,
     month = jan,
     volume = 3,
     pages = {e103},
     journal = {PeerJ Computer Science},
     issn = {2376-5992},
     url = {https://doi.org/10.7717/peerj-cs.103},
     doi = {10.7717/peerj-cs.103}
    }

@article{DBLP:journals/corr/abs-2310-07582,
  publtype={informal},
  author={Dean S. Hazineh and Zechen Zhang and Jeffery Chiu},
  title={Linear Latent World Models in Simple Transformers: A Case Study on Othello-GPT},
  year={2023},
  cdate={1672531200000},
  journal={CoRR},
  volume={abs/2310.07582},
  url={https://doi.org/10.48550/arXiv.2310.07582}
}

@inproceedings{
gurnee2024language,
title={Language Models Represent Space and Time},
author={Wes Gurnee and Max Tegmark},
booktitle={The Twelfth International Conference on Learning Representations},
year={2024},
url={https://openreview.net/forum?id=jE8xbmvFin}
}

@inproceedings{
kuhn2023semantic,
title={Semantic Uncertainty: Linguistic Invariances for Uncertainty Estimation in Natural Language Generation},
author={Lorenz Kuhn and Yarin Gal and Sebastian Farquhar},
booktitle={The Eleventh International Conference on Learning Representations },
year={2023},
url={https://openreview.net/forum?id=VD-AYtP0dve}
}

@InProceedings{Shao_2018_CVPR_Workshops,
author = {Shao, Hang and Kumar, Abhishek and Thomas Fletcher, P.},
title = {The Riemannian Geometry of Deep Generative Models},
booktitle = {Proceedings of the IEEE Conference on Computer Vision and Pattern Recognition (CVPR) Workshops},
month = {June},
year = {2018}
}

@inproceedings{LuBM0S22,
  author={Yao Lu and Max Bartolo and Alastair Moore and Sebastian Riedel and Pontus Stenetorp},
  title={Fantastically Ordered Prompts and Where to Find Them: Overcoming Few-Shot Prompt Order Sensitivity},
  year={2022},
  cdate={1640995200000},
  pages={8086-8098},
  url={https://doi.org/10.18653/v1/2022.acl-long.556},
  booktitle={ACL (1)}
  }

@inproceedings{
Holtzman2020The,
title={The Curious Case of Neural Text Degeneration},
author={Ari Holtzman and Jan Buys and Li Du and Maxwell Forbes and Yejin Choi},
booktitle={International Conference on Learning Representations},
year={2020},
url={https://openreview.net/forum?id=rygGQyrFvH}
}

@inproceedings{
zhang2024how,
title={How Language Model Hallucinations Can Snowball},
author={Muru Zhang and Ofir Press and William Merrill and Alisa Liu and Noah A. Smith},
booktitle={Forty-first International Conference on Machine Learning},
year={2024},
url={https://openreview.net/forum?id=FPlaQyAGHu}
}

@misc{belrose2025elicitinglatentpredictionstransformers,
      title={Eliciting Latent Predictions from Transformers with the Tuned Lens}, 
      author={Nora Belrose and Igor Ostrovsky and Lev McKinney and Zach Furman and Logan Smith and Danny Halawi and Stella Biderman and Jacob Steinhardt},
      year={2025},
      eprint={2303.08112},
      archivePrefix={arXiv},
      primaryClass={cs.LG},
      url={https://arxiv.org/abs/2303.08112}, 
}

@misc{zou2025representationengineeringtopdownapproach,
      title={Representation Engineering: A Top-Down Approach to AI Transparency}, 
      author={Andy Zou and Long Phan and Sarah Chen and James Campbell and Phillip Guo and Richard Ren and Alexander Pan and Xuwang Yin and Mantas Mazeika and Ann-Kathrin Dombrowski and Shashwat Goel and Nathaniel Li and Michael J. Byun and Zifan Wang and Alex Mallen and Steven Basart and Sanmi Koyejo and Dawn Song and Matt Fredrikson and J. Zico Kolter and Dan Hendrycks},
      year={2025},
      eprint={2310.01405},
      archivePrefix={arXiv},
      primaryClass={cs.LG},
      url={https://arxiv.org/abs/2310.01405}, 
}
